\documentclass[final]{neus2025}

\title[Interpretable Generative Adversarial Imitation Learning]{Interpretable Imitation Learning via\\ Generative Adversarial STL Inference and Control
}
\usepackage{times}

\usepackage{appendix}
\usepackage{tikz}
\usetikzlibrary{calc}
\usetikzlibrary{matrix}
\usetikzlibrary{chains}
\usetikzlibrary{shapes.geometric}
\usetikzlibrary{arrows.meta}
\usetikzlibrary{decorations.pathreplacing}
\usetikzlibrary{backgrounds}
\usepackage{xcolor}
\usepackage{amsmath}
\usepackage{amssymb}
\usepackage{graphicx}
\usepackage{url}
\usepackage{algorithm2e}
\usepackage{times}
\usepackage{comment}
\usepackage{caption}
\newtheorem{problem}{Problem}

\def\always{\square}
\def\event{\lozenge}
\definecolor{dodgerblue}{rgb}{0.12, 0.56, 1.0}

\author{
\Name{Wenliang Liu}\footnotemark[1] \Email{wliu97@bu.edu}\\
\Name{Danyang Li} \thanks{These authors contributed equally.} \Email{danyangl@bu.edu}\\
\addr Department of Mechanical Engineering, Boston University, Boston, MA, USA
\AND
\Name{Erfan Aasi} \Email{eaasi@mit.edu}\\
\Name{Daniela Rus} \Email{rus@mit.edu}\\
\addr Computer Science and Artificial Intelligence Laboratory, MIT, Cambridge, MA, USA
\AND
\Name{Roberto Tron} \Email{tron@bu.edu}\\
\addr Department of Mechanical Engineering, Boston University, Boston, MA, USA
\AND
\Name{Calin Belta} \Email{calin@umd.edu}\\
\addr Department of Electrical and Computer Engineering, University of Maryland, College Park, MD, USA
}

\begin{document}
\maketitle
\begin{abstract}
Imitation learning methods have demonstrated considerable success in teaching autonomous systems complex tasks through expert demonstrations. However, a limitation of these methods is their lack of interpretability, particularly in understanding the specific task the learning agent aims to accomplish. In this paper, we propose a novel imitation learning method that combines Signal Temporal Logic (STL) inference and control synthesis, enabling the explicit representation of the task as an STL formula. This approach not only provides a clear understanding of the task but also supports the integration of human knowledge and allows for adaptation to out-of-distribution scenarios by manually adjusting the STL formulas and fine-tuning the policy. We employ a Generative Adversarial Network (GAN)-inspired approach to train both the inference and policy networks, effectively narrowing the gap between expert and learned policies. The efficiency of our algorithm is demonstrated through simulations, showcasing its practical applicability and adaptability.
\end{abstract}


\section{INTRODUCTION}
Imitation learning is a machine learning technique that enables autonomous systems to learn tasks by mimicking expert behavior and is increasingly popular for teaching complex tasks to robotic systems efficiently. This paper focuses on imitation learning using offline data, i.e., no interaction with the expert is required during the learning process. In this setting, imitation learning can be categorized into two main types: behavioral cloning~\cite{pomerleau1991efficient}, which utilizes supervised learning to learn a policy from state-action pairs, and Inverse Reinforcement Learning (IRL)~\cite{ng2000algorithms}, which infers the reward function that the expert optimizes and applies Reinforcement Learning (RL) to find a control policy. Although behavioral cloning is conceptually simple, it mimics the expert without uncovering the underlying task. Hence, it lacks adaptability to situations not covered in the training data. IRL can recover a reward function, but it is computationally expensive, prone to overfitting, and difficult to interpret. Therefore, it is challenging to assess the correctness of the reward function and manipulate it to incorporate human knowledge or adapt to new scenarios. 

Temporal logics, including Linear Temporal Logic (LTL)~\cite{baier2008principles} and Signal Temporal Logic (STL)~\cite{maler2004monitoring}, are widely used for control system specifications due to their expressiveness. STL inference is the process of deriving formal descriptions of system behaviors from observed data in the form of STL formulas~\cite{baharisangari2021uncertainty}. The STL inference approach classifies system behaviors as desired or undesired based on whether they satisfy the inferred formula~\cite{bombara2021offline,aasi2022classification}. In this paper, we propose a method that combines STL inference with control synthesis to enhance the interpretability of imitation learning. Our approach infers an STL formula that describes the task an expert aims to accomplish, and learns a control policy to satisfy this formula in a dynamic environment, where agent dynamics can be either known or learned, and environment dynamics are unknown. Since STL formulas closely resemble natural language, our approach provides some understanding of the objective of the policy. Moreover, they can be manually adjusted to align with expert knowledge or adapt to new conditions. The policy can also be fine-tuned in new scenarios using the learned or modified STL formulas, enabling out-of-distribution generalization without new expert data.

One of the main challenges in combining the STL inference and control tasks is the need to have both positive (expert demonstrations) and negative (incorrect behaviors) examples in order to infer an STL formula classifier. The goal is to push the decision boundary (in the form of an STL formula) as close to the positive data as possible to align the synthesized control policy with the expert’s policy. However, negative examples are often unavailable or insufficient. Although we can record negative data or manually generate them using a simulator, they can hardly be comprehensive enough. To address this concern, we incorporate Generative Adversarial Networks (GANs)~\cite{goodfellow2014generative} into our framework.
We consider the policy network as the generator and the inference network as the discriminator. The policy generates fake (negative) data, and the STL formula distinguishes them from positive data.
We show that by iteratively training the two networks, the learned policy gradually approaches the expert's policy.

The contributions of this paper are as follows: (1) We develop an interpretable imitation learning approach for dynamic environments by integrating STL inference with control synthesis, where the tasks that the expert aims to accomplish are explicitly learned as STL formulas. (2) We incorporate GANs into our framework to gradually bridge the gap between expert and learned policies. (3) We illustrate the efficacy of our algorithm through three case studies. We show that the inferred STL formula can be adjusted to incorporate rules from human knowledge, and the policy can be retrained and adapted to unseen scenarios.

\vspace{-2pt}
\section{Related Work}

STL inference and control synthesis, as separate areas, have received significant attention recently. Early efforts in STL inference focused on mining optimal parameters for predefined formula templates \cite{asarin2012parametric, jin2013mining, jha2017telex, hoxha2018mining}. Recent studies have proposed general learning frameworks to infer both formula structures and their parameters, using techniques such as lattice search~\cite{kong2016temporal}, decision trees~\cite{bombara2021offline,aasi2022classification}, enumeration-based methods~\cite{mohammadinejad2020interpretable}, and neural networks~\cite{chen2022interpretable,li2023learning}. Other method involves using ``landmarks" to build a policy summary~\cite{sreedharan2020tldr}. On the other hand, control synthesis from STL formulas can be solved using mixed integer programming (MIP) \cite{raman2014model,sadraddini2015robust}, or gradient-based optimization~\cite{pant2017smooth,haghighi2019control,gilpin2020smooth}. Recently, learning-based control methods under STL specifications have been proposed, including Q-learning~\cite{aksaray2016q} and model-based methods~\cite{yaghoubi2019worst,liu2021recurrent,leung2022semi,liu2023safe}.
The latter considered static environments, while in this paper we consider a learning agent in a dynamic environment.

Integrating STL inference with control synthesis has gained limited attention in existing literature. One related study \cite{xu2018advisory} learns advice in the form of STL formulas from successful and failed trajectories, and design an advisory controller to satisfy the inferred STL formulas. However, this approach is restricted to a specific template of STL formulas and requires expert intervention during the learning process. In their method, decision trees are used for inference, and the controller is synthesized using a MIP solver. In contrast, our approach enables the template-free learning of STL formulas, and both inference and control synthesis are based on neural networks, leading to a simpler training process and more efficient online execution using offline data only.

The combination of GANs and imitation learning was first proposed as Generative Adversarial Imitation Learning (GAIL) in \cite{ho2016generative} and has been studied extensively in the literature, e.g., \cite{baram2017end, wang2017robust}. Similar to the existing works in imitation learning, these approaches lack interpretability. To the best of our knowledge, this paper is the first to use GANs to integrate temporal logic inference and control synthesis.

\vspace{-2pt}
\section{Problem Statement and Approach}
\label{sec:prob-form}


We consider a discrete-time system consisting of an agent and its environment. The state of the system at time $t\in[0,T]\cap\mathbb Z$ is denoted as $x(t) = \big(x_{ag}(t),x_{env}(t)\big)$, where $x_{ag}(t)\in \mathbb R^{n_a}$ is the state of the agent, $x_{env}(t)\in\mathbb R^{n_e}$ is the state of the environment, and $T\in\mathbb Z^{>0}$ is the time horizon that we are interested in. Let the dynamics of the agent (independent from the environment) be:
\vspace{-6pt}
\begin{equation}
\label{eq:dynamics}
    x_{ag}(t+1) = f\big(x_{ag}(t),u(t)\big),
    \vspace{-6pt}
\end{equation}
where $u(t)\in\mathcal U\subset \mathbb R^m$ is the control input at time $t$, $\mathcal U$ is a set capturing the control constraints. We assume $\mathcal U$ is a box constraint and $f$ is a differentiable function. Let $P_0:\mathcal X_0\rightarrow \mathbb R^{\geq 0}$ be a known probability distribution of the initial state for the agent over a set $\mathcal X_0$ and $P:(\mathbb R^{n_e})^{T+1} \rightarrow \mathbb R^{\geq0}$ be an unknown distribution of the environment trajectory. Let $\mathbf x^{t_1:t_2}=[x(t_1),\ldots,x(t_2)]$ denote a sequence of system states from time $t_1$ to $t_2$ and $\mathbf x=\mathbf x^{0:T}$ denote the whole trajectory. We assume the state $x$ can be fully observed by the agent at all times. The agent model \eqref{eq:dynamics} can be either known or separately learned using the method in \cite{liu2023safe}. We consider both cases in Sec~\ref{sec:exp}. 

Assume we have a dataset $D=\{(\mathbf x^i,l^i)\}_{i=1}^N$ of system trajectories, all of length $T$, where $l^i\in\{1,-1\}$ is the label with $1$ indicating expert demonstration and $-1$ indicating undesired behaviors. We allow the dataset to contain only positive samples. Our goal is to generate an interpretable description of the expert's objective and a control policy that resembles the expert's strategy. This description is captured by an STL formula $\phi$ interpreted over $\mathbf y = [y(0),\ldots,y(T)]$, where $y(t) = h\big(x(t)\big)$ is the feature extracted from the system state and $h:\mathbb R^{n_a+n_e}\rightarrow\mathbb R^{n_y}$ is a known differentiable function. In this paper, we consider a fragment of STL with the syntax:
\vspace{-5pt}
\begin{equation}
\label{eq:stl-frag}
    \phi::=\top \mid \mu \mid \neg \phi \mid \phi_1\land\phi_2 \mid \phi_1\lor\phi_2 \mid \event_{[t_1,t_2]}\phi \mid \always_{[t_1,t_2]}\phi,
    \vspace{-5pt}
\end{equation}
where $\mu$ is a predicate $\mu:=a^\top y(t)\geq b$, $a\in\mathbb R^{n_y}$, $b\in\mathbb R$, and $\phi,\phi_1,\phi_2$ are STL formulas. The Boolean operators $\neg,\land,\lor$ are \emph{negation}, \emph{conjunction} and \emph{disjunction}, and the temporal operators $\event$ and $\always$ represent \emph{eventually} and \emph{always}, respectively. $\event_{[t_1,t_2]}\phi$ is true if $\phi$ is satisfied for at least one time point $t\in[t_1,t_2]\cap\mathbb Z$, while $\always_{[t_1,t_2]}\phi$ is true if $\phi$ is satisfied for all time points $t\in[t_1,t_2]\cap\mathbb Z$. 

The quantitative semantics~\cite{donze2010robust}, also known as robustness, of an STL formula $\phi$, denoted as $r(\mathbf y,\phi)$, is a scalar that measures how strongly the formula is satisfied by a signal $\mathbf y$.
The robustness is sound, which means that $r(\mathbf y, \phi) \geq 0$ if and only if $\phi$ is satisfied by $\mathbf y$. 


Consider a history-dependent policy $u(t) = \pi(\mathbf x^{0:t})$. We assume the agent's policy depends on the environment but does not influence it. Our goal can be formulated as:
\vspace{-4pt}
\begin{problem}
\label{problem:main}
Find an STL formula $\phi$ that classifies the positive and negative data in a set $D$, and find a policy $u(t) = \pi(\mathbf x^{0:t})$ that maximizes $r(\mathbf y, \phi$), i.e., the robustness of $\phi$, where $y(t) = h\big(x(t)\big)$.
\end{problem}

\vspace{-4pt}
We parameterize both the STL robustness function and the policy using neural networks (detailed in Sec.~\ref{sec:nn}), referred to as an inference network $r(\mathbf y,\phi_{\theta_I}) = \mathcal I(\mathbf y;\theta_I)$ and a policy network $u(t) = \pi(\mathbf x^{0:t};\theta_P)$, where $\theta_I$ and $\theta_P$ are the neural network parameters, and $\phi_{\theta_I}$ is an STL formula which can be extracted from the inference network parameters $\theta_I$. Then we divide
Prob.~\ref{problem:main} into two subproblems, the inference problem and the control problem, and formulate them as:

\begin{problem}
    \label{pb:inference}
    [Inference] Given a dataset $D$, find the optimal parameters $\theta_I^*$ for the inference network $\mathcal I(\mathbf y;\theta_I)$ such that the inferred STL formula $\phi_{\theta_{I}}$ minimizes the misclassification rate ($MCR$):
    $\theta_I^* = \arg\min_{\theta_I} MCR(\phi_{\theta_{I}},D)$.
\end{problem}

\begin{problem}
    \label{pb:policy}
    [Control] Given system (\ref{eq:dynamics}) 
    and the inference network $\mathcal I(\mathbf y;\theta_I)$, find the optimal parameters $\theta_P^*$ for the policy network $\pi(\mathbf x^{0:t};\theta_P)$ that maximize the expected robustness of $\phi_{\theta_{I}}$:
    \vspace{-2pt}
    \begin{equation}
    \label{eq:pb2}
    \begin{aligned}
        \theta_P^* = & \arg\max_{\theta_P} \quad \mathbb E_{x_{ag}(0)\sim P_0,\mathbf x_{env}\sim P}\ \mathcal I(\mathbf y;\theta_I)\\
        \text{s.t.} \quad & x_{ag}(t+1) = f\big(x_{ag}(t),\pi(\mathbf x^{0:t};\theta_P)\big),\ y(t) = h(x(t)).
    \end{aligned}
    \vspace{-2pt}
    \end{equation}
\end{problem}

\begin{figure}
    \centering
    \includegraphics[width=0.85\linewidth]{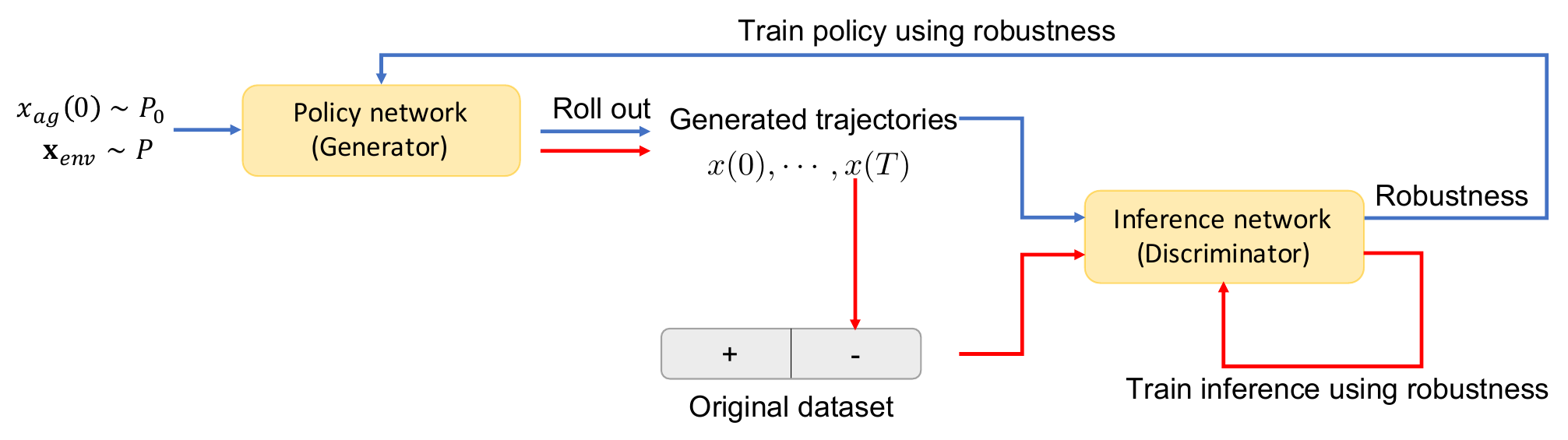}
    \caption{\small An overview of the Generative Adversarial STL Inference and Control: red arrows represent inference network training, while blue arrows indicate policy network training.}
    \label{fig:overview}
\end{figure}

Solving Prob.~\ref{pb:inference} and Prob.~\ref{pb:policy} in one shot may not lead to an accurate STL formula describing the expert's behavior and a policy close enough to the expert's policy. The STL formula $\phi_{\theta_I}$ can be considered as the decision boundary of a binary classifier. In order to accurately explain the underlying task, this decision boundary is expected to be as close as possible to the positive data.  However, the recorded or manually designed negative data can hardly be comprehensive enough to push the decision boundary toward the positive data, because there exists many patterns of undesired behaviors of the system. For instance, it is possible to generate negative data for specific incorrect behaviors, such as a vehicle running a red light, while there are many other undesired behaviors—such as stopping in the middle of the road or frequent acceleration and deceleration. To address this challenge, we propose a generative adversarial training approach. We frame Prob.~\ref{problem:main} as a game between an inference network and a policy network. At each iteration, the policy network generates new trajectories, which are then added to the dataset as negative samples, and the inference network is retrained with this updated data. The policy network aims to generate expert-like trajectories that are difficult for the inference network to classify, while the inference network learns to differentiate between expert and non-expert behaviors. This setup mirrors the roles of the generator and discriminator in a Generative Adversarial Network (GAN), as illustrated in Fig.~\ref{fig:overview}.

\vspace{-2pt}
\section{Neural Network Architectures}
\label{sec:nn}

\subsection{STL Inference Based on Neural Networks}
\label{sec:prelim-inf-net}

In this paper, we apply the TLINet~\cite{li2023learning,li2024tlinet} as the inference network, which is a template-free model that provides flexibility in learning various structures of STL formulas. A TLINet consists of multiple layers, each containing modules corresponding to specific types of operators. These modules are categorized into three types: predicate, temporal and Boolean. The predicate module learns the predicate type and spatial parameters; the temporal module learns the type of temporal operator and temporal parameters; and the Boolean module learns the type of Boolean operator and the structure of the formula. This capability provides flexibility in generating STL formulas, as it is not constrained by a fixed network structure. TLINet allows for flexible combinations of layers, enabling it to describe diverse behaviors. Fig. \ref{fig:TLINet} shows an example of the TLINet structure.

A TLINet can be transferred to an STL formula $\phi_{\theta_I}$ by decoding its parameters $\theta_I$ and translating modules to compositions of $\phi_{\theta_I}$. A detailed decoding and translation example is provided in Appendix~\ref{TLINet}. With a given signal $\mathbf y$ as input, the output of a TLINet is an estimated robustness value $\tilde{r}(\mathbf y, \phi_{\theta_I}) = \mathcal I(\mathbf y;\theta_I)$,
where $\tilde{r}(\mathbf y, \phi_{\theta_I})$ is a smooth approximation of the true robustness. We use Sparse Softmax~\cite{li2024tlinet} and Averaged Max~\cite{li2024tlinet} to approximate the robustness, which allows to incorporate gradient-based methods and can guarantee soundness. Thus, $\tilde{r}(\mathbf y, \phi_{\theta_I})$ is able to represent a rigorous classifier, i.e., $\tilde{r}(\mathbf y, \phi_{\theta_I})$ is positive if and only if the input signal belongs to the positive class and satisfies the STL formula $\phi_{\theta_I}$.

\begin{figure}
\centering
\begin{tikzpicture}
\begin{scope}[every node/.style={minimum width=1.5cm,draw}, align=center]
\node[draw=dodgerblue,text width=1.3cm,font=\fontsize{8}{8}\selectfont] (layer 1a) {\textsf{Predicate Layer}};
\node[draw=black!30!green,right=5mm of layer 1a,text width=1.3cm,font=\fontsize{8}{8}\selectfont] (layer 2a) {\textsf{Boolean Layer}};
\node[draw=orange,right=5mm of layer 2a,text width=1.3cm,font=\fontsize{8}{8}\selectfont] (layer 3a) {\textsf{Temporal Layer}};
\node[draw=orange,right=5mm of layer 3a,text width=1.3cm,font=\fontsize{8}{8}\selectfont] (layer 4a) {\textsf{Temporal Layer}};
\node[draw=black!30!green,right=5mm of layer 4a,text width=1.3cm,font=\fontsize{8}{8}\selectfont] (layer 6) {\textsf{Boolean Layer}};
\end{scope}

\node[circle,fill=orange!70,minimum size=0.8cm] at ($(-1.8,0)$) (s) {$\mathbf{y}$};
\node[circle,fill=orange!70,minimum size=0.8cm, right =5mm of layer 6] (r) {$\tilde{r}$};

\begin{scope}[-latex]
\draw (s) -- (layer 1a);
\draw (layer 1a) -- (layer 2a);
\draw (layer 2a) -- (layer 3a);
\draw (layer 3a) -- (layer 4a);
\draw (layer 4a) -- (layer 6);
\draw (layer 6) -- (r);
\end{scope}
\end{tikzpicture}
\caption{An example of a $5$-layer inference network.}
\label{fig:TLINet}
\end{figure}

\subsection{Recurrent Neural Network Control Policy}
\label{sec:prelim-pol-net}

In general, to satisfy an STL specification, the control policy needs not only the current state but also past states~\cite{liu2021recurrent}. For example, if a robot is required to move back and forth between two regions, then in the middle of the two regions the robot will need the history information of which region it has just visited to decide where to go. Hence, we apply a Recurrent Neural Network (RNN) policy $\pi(\mathbf x^{0:t};\theta_P)$, where $\theta_P$ is the RNN parameters. Similar to \cite{yaghoubi2019worst}, we apply a hyperbolic tangent function on the output of the RNN to satisfy the control constraint $\mathcal U$.

\vspace{-2pt}
\section{Training of Inference and Policy Networks}
\label{sec:training}
Now we elaborate the training processes of the networks, i.e., the solutions to Prob.~\ref{pb:inference} and Prob.~\ref{pb:policy}. Then we detail the joint training of the two networks using the generative adversarial approach.
\subsection{Training of Inference Network}
\label{sec:training-inference}
Following \cite{li2024tlinet}, we solve Prob.~\ref{pb:inference} by minimizing the following loss function:
\vspace{-4pt}
\begin{equation}
    \label{eq:inference}
\begin{split}
    \theta_I^*, \epsilon^* = \arg\min_{\theta_I,\epsilon} &\frac{1}{N}\sum_{i=1}^N ReLU\big(\epsilon - l^i\cdot \mathcal{I}(\mathbf y^i, \theta_I)\big) - \beta_1 \epsilon + \beta_2 Reg(\theta_I), 
\end{split}
\end{equation}
where $\epsilon>0$ is the margin, $Reg$ is a regularizer to adjust formula complexity, which is a combination of a bi-modal regularizer~\cite{Murray2010AnAF} as well the traditional $\ell 2$ regularizer. $\beta_1>0, \beta_2\in\mathbb{R}$ are two hyperparameters to control the compromise among formula complexity, maximizing the margin $\epsilon$, and minimizing the $MCR$. The loss function \eqref{eq:inference} satisfies that it is small when the inferred formula is satisfied by the positive data and violated by the negative data, and it is large otherwise. The margin $\epsilon$ is a quantitative measurement of the separation of signals by $\phi_{\theta_I}$. The loss function encourages a large margin to better distinguish the positive and negative data.

\subsection{Training of Policy Network}
\label{sec:training-policy}
Model-based training of an RNN control policy in a static environment was introduced in \cite{liu2023safe}. In this paper, we extend this method to dynamic environments. To solve Prob.~\ref{pb:policy}, we need to estimate the expectation in \eqref{eq:pb2}. We first sample a set of $M$ initial agent states $\{\bar x_{ag}^{j}(0)\}_{j=1}^{M}$ from the known distribution $P_0$. Next, since the distribution $P$ is unknown, we sample $M$ environment trajectories $\{\bar{\mathbf x}_{env}^{j}\}_{j=1}^{M}$ from the dataset $D$. We use the agent dynamics \eqref{eq:dynamics}, policy network $\pi$, and the sampled environment trajectories to roll out the system trajectories from the sampled initial agent states. Then we use the mean value to estimate the expectation, attaining:
\vspace{-2pt}
\begin{equation}
\label{eq:policy_train}
    \begin{aligned}
        \theta_P^* = & \arg\max_{\theta_P} \quad  \frac{1}{M}\sum_{j=1}^M \mathcal I\big(\bar{\mathbf y}^i;\theta_I\big)\\
        & \text{s.t.} \quad  \bar x_{ag}^j(t+1) = f\big(\bar x_{ag}^j(t),\pi(\bar{\mathbf x}^{0:t,j};\theta_P)\big),\ \bar y^j(t) = h\big(\bar x^j(t)\big),
    \end{aligned}
\end{equation}
where the closed-loop dynamics can be substituted into the objective to form an unconstrained optimization problem. Since both inference $\mathcal I$ and policy $\pi$ are based on neural networks, and the dynamics \eqref{eq:dynamics} are differentiable, the gradient of the objective can be backpropagated to the policy parameters. At each optimization step, we resample the initial states and environment trajectories to get an unbiased estimation of the objective and optimize it using Adam~\cite{kingma2014adam}.

\subsection{Generative Adversarial Training}
\label{sec:training-ga}
In this section, we describe the generative adversarial training approach used to iteratively refine both the inference and policy networks to improve the overall solution. We begin by training the inference network, followed by training the policy network. Next, we sample a set of initial states and environment trajectories, apply the learned policy, and roll out system trajectories, which are assigned negative labels and added to the dataset. This process (called an iteration) is repeated until convergence. If the original dataset contains only positive data, we first generate negative-labeled trajectories using a random policy and follow the same procedure. To improve the stability of training, we balance the positive and negative samples by randomly selecting an equal number of negative samples as expert demonstrations from the pool of collected negative data at each iteration.

During training, as the policy network improves, trajectories resembling the expert demonstrations are added to the negative dataset. This can cause an increase in the $MCR$ of the inferred STL formula, which may, in turn, degrade the performance of the policy network. To mitigate this issue, we introduce a performance score based on Dynamic Time Warping (DTW)~\cite{sakoe1978dynamic} to assess the control policy. Specifically, we apply the learned policy $\pi$ to the same initial conditions and environment trajectories as those used in the expert demonstrations and compute the similarities between the generated trajectories $\mathbf x^{i,\pi}$ and the expert demonstrations $\mathbf x^i$, which gives us a performance score $r(\pi)$ for the policy $\pi$: 
\vspace{-2pt}
\begin{equation}
\begin{aligned}
    &r(\pi) =  -\sum_{\{i|l_i=1\}}DTW(\mathbf x^{i,\pi},\mathbf x^i),\ \text{where}\\ 
    &\mathbf x^{i,\pi}(0) = \mathbf x^i(0),\ x^i_{env}(t) = x^{i,\pi}_{env}(t),  \ x^{i,\pi}_{ag}(t+1) = f\big(x^{i,\pi}_{ag}(t),\pi(\mathbf x^{i,\pi}_{0:t};\theta_P)\big).
\end{aligned}
\end{equation}
We select the policy with the highest performance score as the final policy, with the corresponding STL formula as its interpretation. 

\section{Experiments}
\label{sec:exp}

In this section, we evaluate our algorithm through three case studies. The first one involves a navigation task with random goal and obstacles. We compare our algorithm with Behavioral Cloning (BC) and show out-of-distribution generalization of our approach. In the second, we apply our approach to tasks in the MuJoCo environments with unknown dynamics, and compare its performance with GAIL. The third one includes a self-driving vehicle in a dynamic environment. We show how user-generated formulas can be integrated into the policy to incorporate human knowledge and how the policy can be efficiently adapted to previously unseen scenarios through formula modification. Some detailed setup and results of experiments are omitted here and given in Appendix \ref{appendix}. Videos of the experiments can be found at \url{https://youtu.be/peQSizAwNAo}. The implementation is available at: \url{https://github.com/danyangl6/IGAIL}.

\subsection{Random Navigation}\label{sec:navigation}
\begin{figure}
    \centering
    \subfigure[]{
    \centering
    \includegraphics[width=0.46\textwidth]{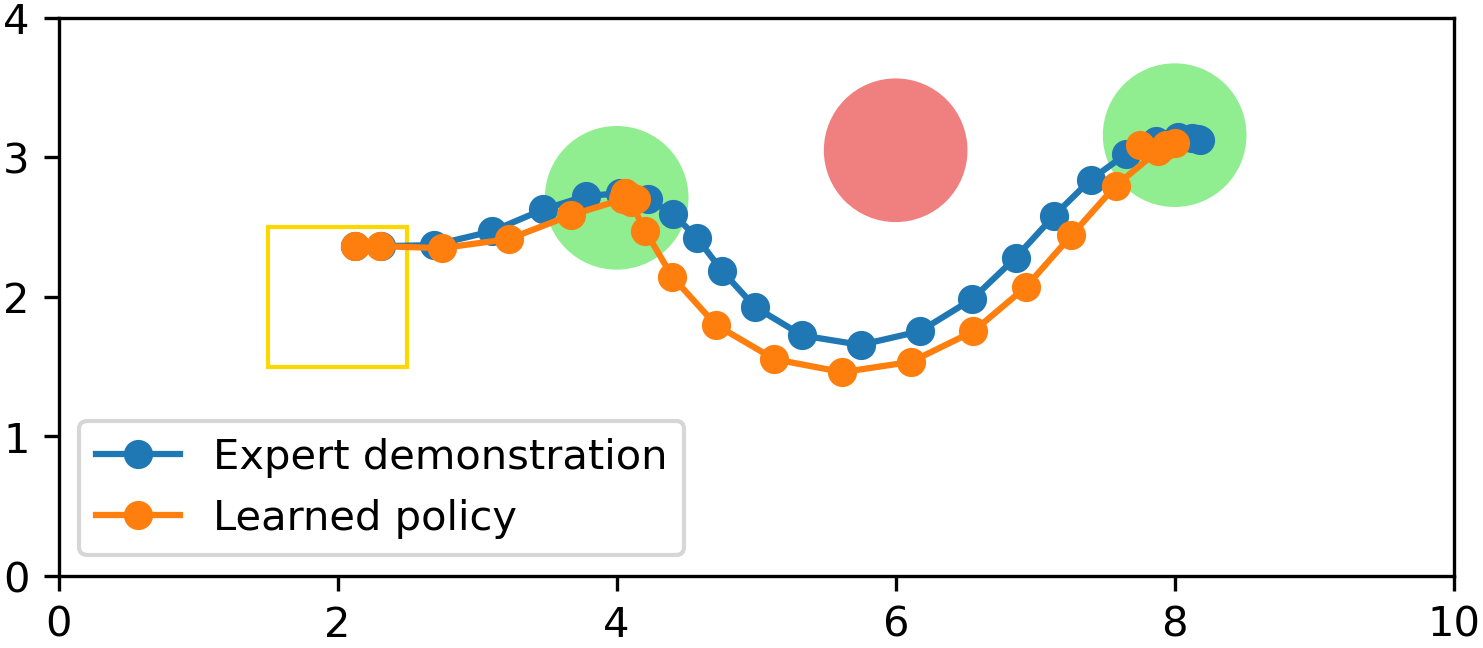}
        \label{fig:navigation}
    }
    \subfigure[]{
        \centering
        \includegraphics[width=0.46\textwidth]{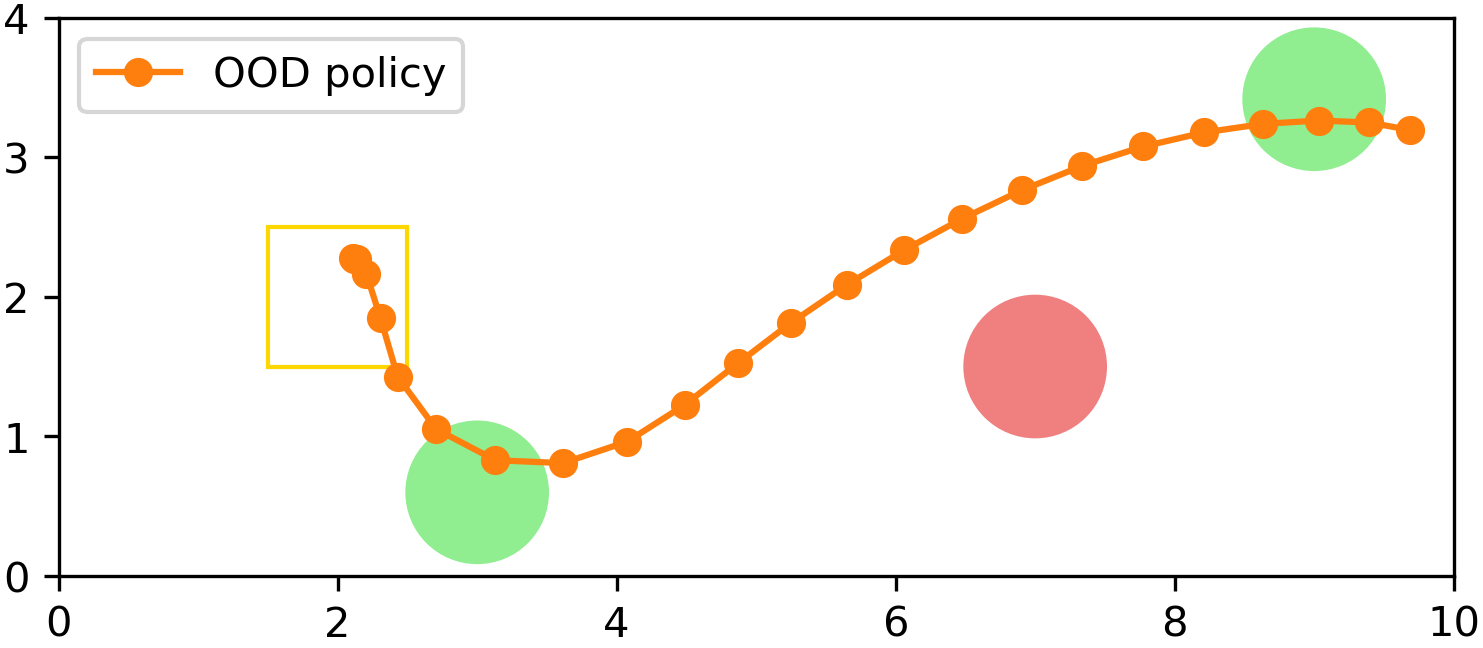}
        \label{fig:ood}
    }
    \caption{(a) Expert demonstration and sampled trajectory using the learned policy. (b) Sampled trajectory using the learned policy after fine-tuning in an OOD environment.}
    \label{fig:unicycle}
\end{figure}

Consider a unicycle robot moving in a 2D environment with known discrete-time dynamics as shown in Fig~\ref{fig:unicycle}. The state of the robot $x_{ag} = (p_x,p_y,\theta)$ contains its 2D position $p_x$, $p_y$ and orientation $\theta$, and the control $u=(v,\omega)$ includes the forward velocity $v$ and angular velocity $\omega$. The initial state is uniformly sampled from $[1.5,2.5]^2\times[-0.01,0.01]$. The environment contains two goal regions and one obstacle region, each with a radius of $0.5$. They are centered at random $y$-coordinates uniformly sampled from $[0.5,3.5]$ (with fixed $x$-coordinates). The robot must first visit the closer goal region, then the other one, and always avoid the obstacle, with $p_y\in[0,4]$. This behavior is specified by an STL formula (unknown to the learner), which is provided in Appendix~\ref{appendix}. The environment state includes the static locations of the goal and obstacle regions, i.e., $x_{env}(t) = x_{env}(0)$, $\forall t\in[0,T]$. We collect a dataset of $1000$ positive robot trajectories with a time horizon $T=24$ using an expert policy trained through model-based policy search~\cite{liu2023safe}.

\begin{table}[]
    \centering
    \begin{tabular}{l c c c c c}
        \hline
              & Ours & BC & Expert & Ours (OOD) & BC (OOD)  \\
        \hline
            Robustness & $0.19\pm0.13$ & $0.31\pm0.08$ & $0.33\pm0.06$ & $0.12\pm0.16$ & $-0.12\pm0.13$ \\
            Suc. rate & $85.4\pm14.5\%$ & $97.5\%$ & $99.2\%$ & $72.7\%\pm22.7\%$ & $17.1\%$\\
        \hline
        \end{tabular}
    \caption{Robustness and success rate (mean$\pm$std) of the learned policy w.r.t. the true STL formula}
    \label{tb:navigation}
\end{table}

We evaluate our algorithm from $6$ random initializations of both networks. The learned STL formula (given in Appendix~\ref{appendix}) closely reproduces the specification the expert aims to satisfy. Fig.~\ref{fig:navigation} shows a sampled trajectory generated using the learned policy starting from the same initial state as the expert demonstration. Table.~\ref{tb:navigation} lists the robustness of the learned policy with respect to the ground-truth STL formula and the success rate of satisfying it, evaluated over $1000$ random initial states for each network initialization. We compare the learned policy with the expert policy and BC policy trained using the same set of expert demonstrations. Our policy achieves relatively high robustness and success rate. Although its performance does not match that of BC, it provides a reasonable description of the task using the STL formula. This interpretability is absent in BC.

To demonstrate the out-of-distribution (OOD) generalization of our approach, we modify the environment by changing the $x$ coordinates of the $3$ regions ($y$ coordinates are still random) as shown in Fig.~\ref{fig:ood}. Without any new expert data, we fine-tune the policy network in this new environment with the same inferred STL formula. We compare the performance with the BC policy, which cannot be retrained due to the lack of available data. The statistics in Table.~\ref{tb:navigation} show that after fine-tuning, our policy maintains relatively high robustness and success rate, while the performance of the BC policy dropped significantly.


\subsection{MuJoCo Tasks}\label{sec:mujoco}

We evaluate our algorithm against baselines on $2$ control tasks, the Inverted Pendulum with $x\in\mathbb R^4$ and the Reacher with $x\in \mathbb R^{11}$, as shown in Fig.~\ref{fig:mujoco} (detailed in Appendix~\ref{appendix}).  All environments are simulated with MuJoCo~\cite{todorov2012mujoco} and the system dynamics are unknown and learned together with the policy as in \cite{liu2023safe}. Each task comes with a true reward function, defined in the OpenAI Gym~\cite{brockman2016openai}. As modern RL approaches, such as PPO~\cite{schulman2017proximal}, can generate high-quality policies for these tasks, we first run PPO on each task to create $1000$ expert trajectories, each trajectory is truncated or padded to $50$ steps. We test our algorithm on this dataset and compare it with the standard GAIL~\cite{ho2016generative}. Learned STL formulas (given in Appendix~\ref{appendix}) effectively describe the tasks in an interpretable manner. The rewards gained by our approach, GAIL, and the expert policy are given in Table~\ref{tb:mujoco}. The statistics of our algorithm are computed over $5$ network initializations, with $1000$ trajectories generated for each initialization. 

\begin{figure}
\centering
\begin{minipage}{0.3\linewidth}
\centering
\includegraphics[width=\textwidth]{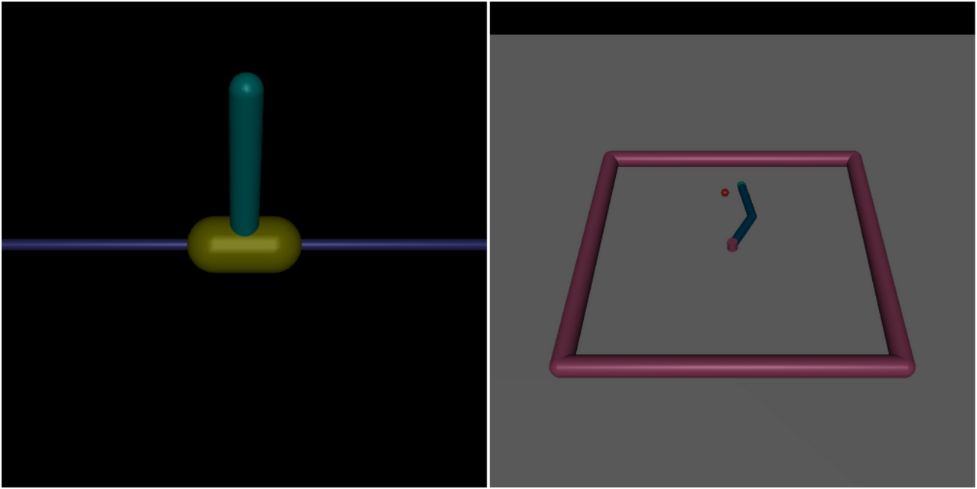}
\captionof{figure}{\small Inverted Pendulum (left) and Reacher (right).}
\label{fig:mujoco}
\end{minipage}
\hfill
\begin{minipage}{0.65\linewidth}
\centering
\begin{tabular}{cccc}
    \hline
    Task &Ours &GAIL &Expert \\
    \hline
    Pendulum & $50.0\pm0.0$ & $50.0\pm0.0$ & $50.0\pm0.0$ \\
    Reacher & $-4.76 \pm 1.94$ & $-5.04 \pm 2.22$ & $-4.31\pm 1.65$ \\
    \hline
    \end{tabular}
\captionof{table}{Rewards (mean$\pm$std) obtained by different polices.}
\label{tb:mujoco}
\end{minipage}
\vspace{-10pt}
\end{figure}

For the Inverted Pendulum task, both our approach and GAIL reach the maximum reward achievable in 50 steps. For the Reacher task, our approach slightly outperforms GAIL. However, for MuJoCo tasks such as Swimmer and Walker, which are hard to express by the STL fragment in \eqref{eq:stl-frag}, our approach does not reach good reward levels. In future work, we plan to investigate ways to enhance the expressivity of the inference network to address this limitation.

\subsection{A Self-Driving Scenario}\label{sec:case-driving}

\begin{figure}
    \centering
    \subfigure[]{
        \includegraphics[scale=0.23]{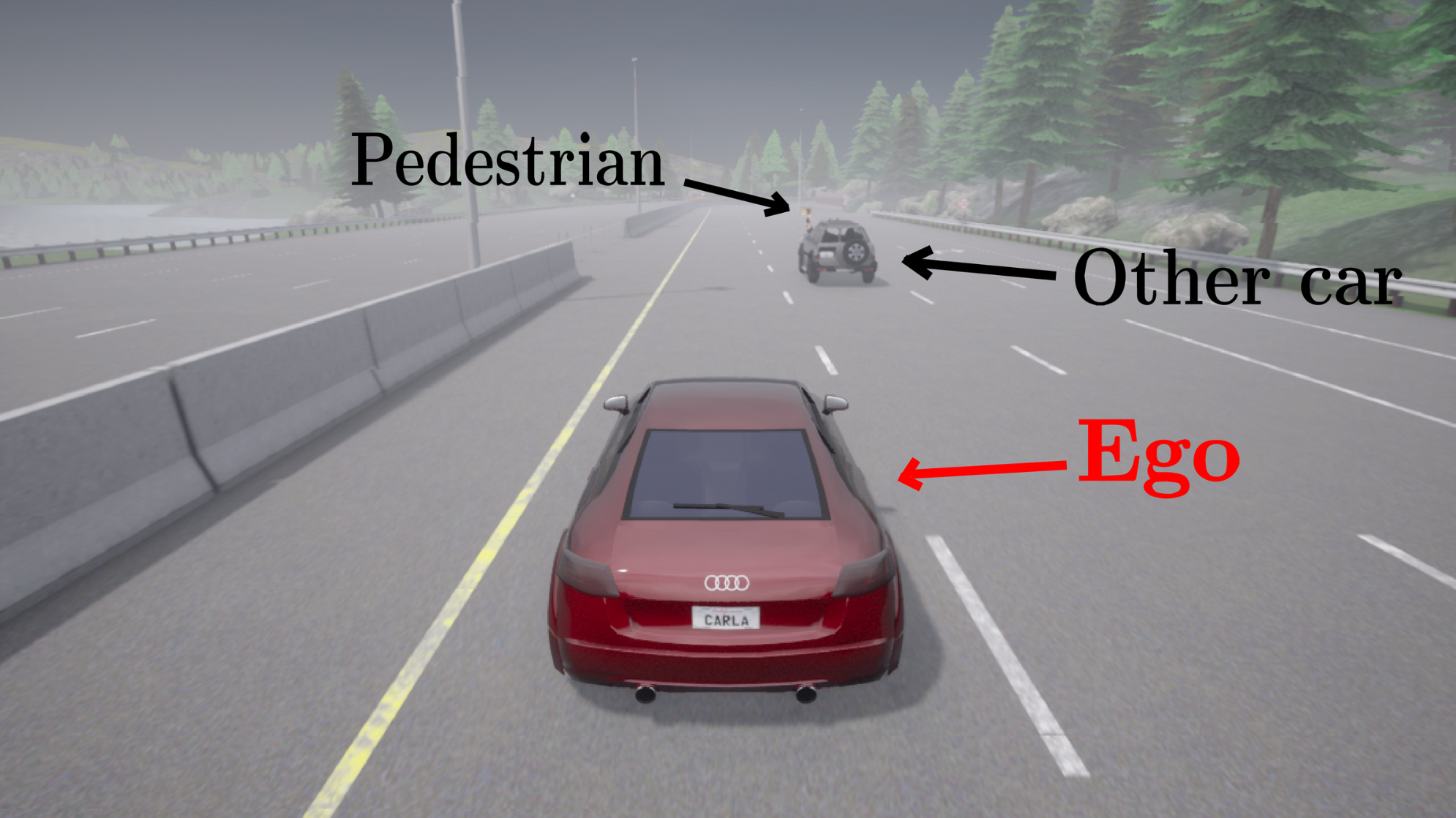}
        \label{fig:carla}
    }
    \ 
    \subfigure[]{
        \includegraphics[width=.48\textwidth]{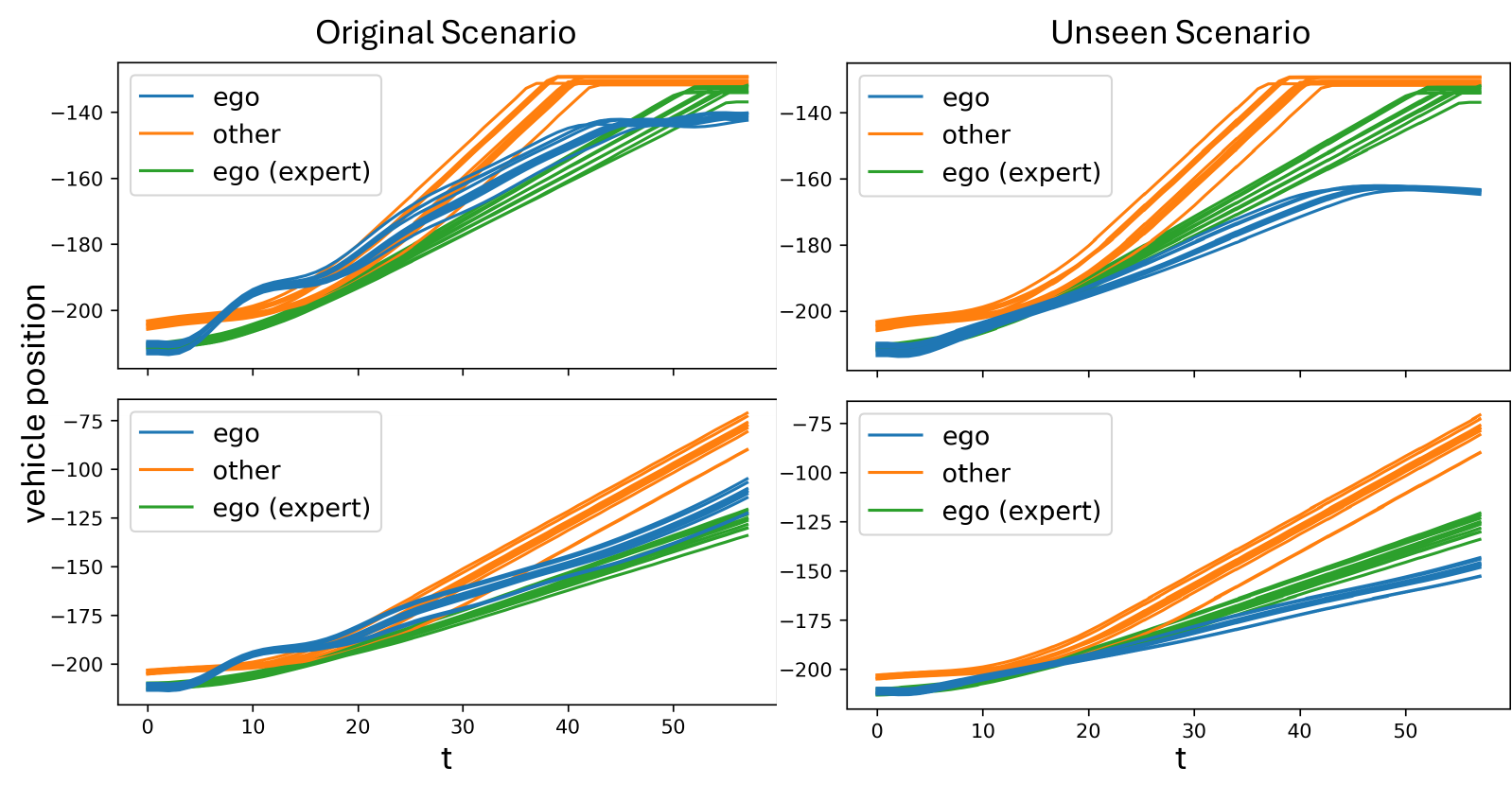}
        \label{fig:traj}
    }
    \caption{\small (a) The autonomous driving scenario simulated in Carla. (b) Sampled trajectories of ego and other vehicles' positions after training in the original and unseen scenarios. The top figures show the situations when there are pedestrians, and the bottom figures show the situations when there is no pedestrian.}
    \label{fig:case2}
    \vspace{-10pt}
\end{figure}

Inspired by \cite{aasi2022classification}, we consider the following scenario: a self-driving vehicle (ego) drives in an urban environment alongside another vehicle driven by a ``reasonable" human in the adjacent lane. The initial position of the other vehicle is always ahead of the ego, and both are headed toward an unmarked crosswalk. If a pedestrian crosses, the other vehicle brakes to stop; otherwise, it continues moving without deceleration. Due to foggy weather and the other vehicle obstructing the view, ego needs to infer the presence of a pedestrian by observing the behavior of other vehicle (see Fig.~\ref{fig:carla}). Under the assumption that ego only moves forward, we simplify its dynamics as a double integrator (assumed to be known). The system state $x(t)\in\mathbb R^4$ is the concatenation of the agent state $x_{ag}(t) = \big(p_{eg}(t),v_{eg}(t)\big)$ and the environment state $x_{env}(t) = \big(p_{ot}(t),v_{ot}(t)\big)$, where $p_{eg}(t)$ and $v_{eg}(t)$ are the position and velocity of ego; $p_{ot}(t)$ and $v_{ot}(t)$ are the position and velocity of the other vehicle. The control $u(t)$ is the same as ego's acceleration $a_{eg}(t)\in\mathcal U\subset \mathbb R$. We use the autonomous driving simulator Carla~\cite{Dosovitskiy17} to generate a dataset of $800$ trajectories, containing both positive and negative labels, and they both include situations with and without pedestrians ($200$ data for each situation). The time horizon $T=57$.

\textbf{Incorporate known critical rules:} Some critical rules from human knowledge can be manually incorporated as supplementary formulas into the inferred STL formula to make the policy satisfy them. Here, we formulate the speed limit rule as $\phi_{sl}:= \always_{[0,57]} \big((v_{eg}\leq 10)\land (v_{eg}> -1)\big)$. We encode $\phi_{sl}$ into the inference network as a conjunction to the inferred formula: $\tilde{\phi}_{\theta_I}=\phi_{\theta_I}\land\phi_{sl}$. We test the learned policy in Carla and track the output of the policy. Sampled trajectories are shown in Fig.~\ref{fig:traj} and the inferred formula is provided in Appendix \ref{appendix}, which clearly reveals the logic the ego should obey. Since we apply a more complex inference network structure in this case, we use dual annealing~\cite{xiang1997generalized}, a global optimizer to find one good initialization of $\theta_I$ and then run our algorithm. In this case, we do not have a ground-truth STL formula or reward function, so we do not compute statistics of the learned policy. But as shown in Fig.~\ref{fig:traj}, the ego vehicle successfully stops in front of the pedestrian when it detects the deceleration of the other vehicle. Otherwise, it continues driving without slowing down.

\textbf{Unseen scenarios:} We can adapt the policy to similar but unseen scenarios by tuning the inferred formula. Assume that the speed limit is restricted to $4$ due to road construction. This is an unseen scenario and both the other and the expert-controlled ego vehicles violate the new rule. We modify the formula $\phi_{sl}$ by replacing $v_{eg}\leq 10$ with $v_{eg}\leq 4$ and retrain the policy network. Sampled trajectories, shown in Fig.~\ref{fig:traj}, indicate that the new policy successfully adheres to the updated speed limit while maintaining compliance with previous rules.

\section{Discussion}
As demonstrated, our method can generate a control policy while simultaneously providing a task description using STL formulas. However, this interpretability comes with trade-offs in performance for several reasons. First, the inference network is not a universal classifier, making it less suitable for tasks that are difficult to describe with an STL formula, such as the Swimmer and Walker tasks in MuJoCo. Our approach is better suited for imitating expert demonstrations under spatial-temporal requirements. 
Second, the inference network requires more expert demonstrations compared to traditional imitation learning. Small dataset can result in a conservative, under-approximated STL formula. 
Finally, the policy is trained based on the information provided by the inference network. When negative data become more and more similar to expert demonstrations, 
the inferred formula may degrade, which negatively impacts the policy’s effectiveness. Consequently, a performance score is needed to select the most suitable policy, though our observations suggest that the policy chosen by this score is not always the best one in the training process. We believe that performance could be further improved with a more robust selection criterion.

Our approach prioritizes interpretability, offering insight into the objectives of the expert rather than simply mimicking its behavior. This interpretability enables the integration of safety constraints as supplementary STL formulas. Additionally, it offers flexibility and adaptability; once the task is described, the policy can be fine-tuned in a new but similar environment with the learned or manually adjusted STL formula. This is highly sample-efficient, as there is no need to generate new expert demonstrations, which is typically a challenging aspect of imitation learning.

\section{Conclusion and Future Work}
We proposed an interpretable imitation learning method that combines STL inference and control synthesis in a generative adversarial manner. We use three case studies to demonstrate that our method can uncover the underlying rules from the expert demonstrations in the form of an STL formula, and learn the policy to satisfy these rules. We also showed that we can manually add and adjust rules to adapt the policy for unseen scenarios. Future work includes improving the computational efficiency and increasing the expressiveness of the inference network.

\acks{Partial support for this work was provided by the NSF under grant 2422282 and by the AFOSR under grant FA9550-23-1-0529. This work was also supported by Toyota Research Institute (TRI). It, however, reflects solely the opinions and conclusions of its authors, and not TRI or any other Toyota entity. This work was also partly supported by the USARMY Integrated Fire Control Technology program under Air Force Contract No. FA8702-15-D-0001. Any opinions, findings, conclusions, or recommendations expressed in this publication are those of the authors and don’t necessarily reflect the views of the sponsors.}

\bibliography{references}

\begin{thebibliography}{42}
\providecommand{\natexlab}[1]{#1}
\providecommand{\url}[1]{\texttt{#1}}
\expandafter\ifx\csname urlstyle\endcsname\relax
  \providecommand{\doi}[1]{doi: #1}\else
  \providecommand{\doi}{doi: \begingroup \urlstyle{rm}\Url}\fi

\bibitem[Aasi et~al.(2022)Aasi, Vasile, Bahreinian, and Belta]{aasi2022classification}
Erfan Aasi, Cristian~Ioan Vasile, Mahroo Bahreinian, and Calin Belta.
\newblock Classification of time-series data using boosted decision trees.
\newblock In \emph{2022 IEEE/RSJ International Conference on Intelligent Robots and Systems (IROS)}, pages 1263--1268. IEEE, 2022.

\bibitem[Aksaray et~al.(2016)Aksaray, Jones, Kong, Schwager, and Belta]{aksaray2016q}
Derya Aksaray, Austin Jones, Zhaodan Kong, Mac Schwager, and Calin Belta.
\newblock Q-learning for robust satisfaction of signal temporal logic specifications.
\newblock In \emph{2016 IEEE 55th Conference on Decision and Control (CDC)}, pages 6565--6570. IEEE, 2016.

\bibitem[Asarin et~al.(2012)Asarin, Donz{\'e}, Maler, and Nickovic]{asarin2012parametric}
Eugene Asarin, Alexandre Donz{\'e}, Oded Maler, and Dejan Nickovic.
\newblock Parametric identification of temporal properties.
\newblock In \emph{Runtime Verification: Second International Conference, RV 2011, San Francisco, CA, USA, September 27-30, 2011, Revised Selected Papers 2}, pages 147--160. Springer, 2012.

\bibitem[Baharisangari et~al.(2021)Baharisangari, Gaglione, Neider, Topcu, and Xu]{baharisangari2021uncertainty}
Nasim Baharisangari, Jean-Rapha{\"e}l Gaglione, Daniel Neider, Ufuk Topcu, and Zhe Xu.
\newblock Uncertainty-aware signal temporal logic inference.
\newblock In \emph{International Workshop on Numerical Software Verification}, pages 61--85. Springer, 2021.

\bibitem[Baier and Katoen(2008)]{baier2008principles}
Christel Baier and Joost-Pieter Katoen.
\newblock \emph{Principles of model checking}.
\newblock MIT press, 2008.

\bibitem[Baram et~al.(2017)Baram, Anschel, Caspi, and Mannor]{baram2017end}
Nir Baram, Oron Anschel, Itai Caspi, and Shie Mannor.
\newblock End-to-end differentiable adversarial imitation learning.
\newblock In \emph{International Conference on Machine Learning}, pages 390--399. PMLR, 2017.

\bibitem[Bombara and Belta(2021)]{bombara2021offline}
Giuseppe Bombara and Calin Belta.
\newblock Offline and online learning of signal temporal logic formulae using decision trees.
\newblock \emph{ACM Transactions on Cyber-Physical Systems}, 5\penalty0 (3):\penalty0 1--23, 2021.

\bibitem[Brockman(2016)]{brockman2016openai}
G~Brockman.
\newblock Openai gym.
\newblock \emph{arXiv preprint arXiv:1606.01540}, 2016.

\bibitem[Chen et~al.(2022)Chen, Lu, Su, and Kong]{chen2022interpretable}
Gang Chen, Yu~Lu, Rong Su, and Zhaodan Kong.
\newblock Interpretable fault diagnosis of rolling element bearings with temporal logic neural network, 2022.

\bibitem[Donz{\'e} and Maler(2010)]{donze2010robust}
Alexandre Donz{\'e} and Oded Maler.
\newblock Robust satisfaction of temporal logic over real-valued signals.
\newblock In \emph{International Conference on Formal Modeling and Analysis of Timed Systems}, pages 92--106. Springer, 2010.

\bibitem[Dosovitskiy et~al.(2017)Dosovitskiy, Ros, Codevilla, Lopez, and Koltun]{Dosovitskiy17}
Alexey Dosovitskiy, German Ros, Felipe Codevilla, Antonio Lopez, and Vladlen Koltun.
\newblock {CARLA}: {An} open urban driving simulator.
\newblock In \emph{Proceedings of the 1st Annual Conference on Robot Learning}, pages 1--16, 2017.

\bibitem[Gilpin et~al.(2020)Gilpin, Kurtz, and Lin]{gilpin2020smooth}
Yann Gilpin, Vince Kurtz, and Hai Lin.
\newblock A smooth robustness measure of signal temporal logic for symbolic control.
\newblock \emph{IEEE Control Systems Letters}, 5\penalty0 (1):\penalty0 241--246, 2020.

\bibitem[Goodfellow et~al.(2014)Goodfellow, Pouget-Abadie, Mirza, Xu, Warde-Farley, Ozair, Courville, and Bengio]{goodfellow2014generative}
Ian Goodfellow, Jean Pouget-Abadie, Mehdi Mirza, Bing Xu, David Warde-Farley, Sherjil Ozair, Aaron Courville, and Yoshua Bengio.
\newblock Generative adversarial nets.
\newblock \emph{Advances in neural information processing systems}, 27, 2014.

\bibitem[Haghighi et~al.(2019)Haghighi, Mehdipour, Bartocci, and Belta]{haghighi2019control}
Iman Haghighi, Noushin Mehdipour, Ezio Bartocci, and Calin Belta.
\newblock Control from signal temporal logic specifications with smooth cumulative quantitative semantics.
\newblock In \emph{2019 IEEE 58th Conference on Decision and Control (CDC)}, pages 4361--4366. IEEE, 2019.

\bibitem[Ho and Ermon(2016)]{ho2016generative}
Jonathan Ho and Stefano Ermon.
\newblock Generative adversarial imitation learning.
\newblock \emph{Advances in neural information processing systems}, 29, 2016.

\bibitem[Hochreiter(1997)]{hochreiter1997long}
S~Hochreiter.
\newblock Long short-term memory.
\newblock \emph{Neural Computation MIT-Press}, 1997.

\bibitem[Hoxha et~al.(2018)Hoxha, Dokhanchi, and Fainekos]{hoxha2018mining}
Bardh Hoxha, Adel Dokhanchi, and Georgios Fainekos.
\newblock Mining parametric temporal logic properties in model-based design for cyber-physical systems.
\newblock \emph{International Journal on Software Tools for Technology Transfer}, 20:\penalty0 79--93, 2018.

\bibitem[Jha et~al.(2017)Jha, Tiwari, Seshia, Sahai, and Shankar]{jha2017telex}
Susmit Jha, Ashish Tiwari, Sanjit~A Seshia, Tuhin Sahai, and Natarajan Shankar.
\newblock Telex: Passive stl learning using only positive examples.
\newblock In \emph{International Conference on Runtime Verification}, pages 208--224. Springer, 2017.

\bibitem[Jin et~al.(2013)Jin, Donz{\'e}, Deshmukh, and Seshia]{jin2013mining}
Xiaoqing Jin, Alexandre Donz{\'e}, Jyotirmoy~V Deshmukh, and Sanjit~A Seshia.
\newblock Mining requirements from closed-loop control models.
\newblock In \emph{Proceedings of the 16th international conference on Hybrid systems: computation and control}, pages 43--52, 2013.

\bibitem[Kingma and Ba(2014)]{kingma2014adam}
Diederik~P Kingma and Jimmy Ba.
\newblock Adam: A method for stochastic optimization.
\newblock \emph{arXiv preprint arXiv:1412.6980}, 2014.

\bibitem[Kong et~al.(2016)Kong, Jones, and Belta]{kong2016temporal}
Zhaodan Kong, Austin Jones, and Calin Belta.
\newblock Temporal logics for learning and detection of anomalous behavior.
\newblock \emph{IEEE Transactions on Automatic Control}, 62\penalty0 (3):\penalty0 1210--1222, 2016.

\bibitem[Leung and Pavone(2022)]{leung2022semi}
Karen Leung and Marco Pavone.
\newblock Semi-supervised trajectory-feedback controller synthesis for signal temporal logic specifications.
\newblock In \emph{2022 American Control Conference (ACC)}, pages 178--185. IEEE, 2022.

\bibitem[Li et~al.(2023)Li, Cai, Vasile, and Tron]{li2023learning}
Danyang Li, Mingyu Cai, Cristian-Ioan Vasile, and Roberto Tron.
\newblock Learning signal temporal logic through neural network for interpretable classification.
\newblock In \emph{2023 American Control Conference (ACC)}, pages 1907--1914. IEEE, 2023.

\bibitem[Li et~al.(2024)Li, Cai, Vasile, and Tron]{li2024tlinet}
Danyang Li, Mingyu Cai, Cristian-Ioan Vasile, and Roberto Tron.
\newblock Tlinet: Differentiable neural network temporal logic inference.
\newblock \emph{arXiv preprint arXiv:2405.06670}, 2024.

\bibitem[Liu et~al.(2021)Liu, Mehdipour, and Belta]{liu2021recurrent}
Wenliang Liu, Noushin Mehdipour, and Calin Belta.
\newblock Recurrent neural network controllers for signal temporal logic specifications subject to safety constraints.
\newblock \emph{IEEE Control Systems Letters}, 6:\penalty0 91--96, 2021.

\bibitem[Liu et~al.(2023)Liu, Nishioka, and Belta]{liu2023safe}
Wenliang Liu, Mirai Nishioka, and Calin Belta.
\newblock Safe model-based control from signal temporal logic specifications using recurrent neural networks.
\newblock In \emph{2023 IEEE International Conference on Robotics and Automation (ICRA)}, pages 12416--12422. IEEE, 2023.

\bibitem[Maler and Nickovic(2004)]{maler2004monitoring}
Oded Maler and Dejan Nickovic.
\newblock Monitoring temporal properties of continuous signals.
\newblock In \emph{International Symposium on Formal Techniques in Real-Time and Fault-Tolerant Systems}, pages 152--166. Springer, 2004.

\bibitem[Mohammadinejad et~al.(2020)Mohammadinejad, Deshmukh, Puranic, Vazquez-Chanlatte, and Donz{\'e}]{mohammadinejad2020interpretable}
Sara Mohammadinejad, Jyotirmoy~V Deshmukh, Aniruddh~G Puranic, Marcell Vazquez-Chanlatte, and Alexandre Donz{\'e}.
\newblock Interpretable classification of time-series data using efficient enumerative techniques.
\newblock In \emph{Proceedings of the 23rd International Conference on Hybrid Systems: Computation and Control}, pages 1--10, 2020.

\bibitem[Murray and Ng(2010)]{Murray2010AnAF}
Walter Murray and Kien~Ming Ng.
\newblock An algorithm for nonlinear optimization problems with binary variables.
\newblock \emph{Computational Optimization and Applications}, 47:\penalty0 257--288, 2010.
\newblock URL \url{https://api.semanticscholar.org/CorpusID:28009541}.

\bibitem[Ng et~al.(2000)Ng, Russell, et~al.]{ng2000algorithms}
Andrew~Y Ng, Stuart Russell, et~al.
\newblock Algorithms for inverse reinforcement learning.
\newblock In \emph{Icml}, volume~1, page~2, 2000.

\bibitem[Pant et~al.(2017)Pant, Abbas, and Mangharam]{pant2017smooth}
Yash~Vardhan Pant, Houssam Abbas, and Rahul Mangharam.
\newblock Smooth operator: Control using the smooth robustness of temporal logic.
\newblock In \emph{2017 IEEE Conference on Control Technology and Applications (CCTA)}, pages 1235--1240. IEEE, 2017.

\bibitem[Pomerleau(1991)]{pomerleau1991efficient}
Dean~A Pomerleau.
\newblock Efficient training of artificial neural networks for autonomous navigation.
\newblock \emph{Neural computation}, 3\penalty0 (1):\penalty0 88--97, 1991.

\bibitem[Raman et~al.(2014)Raman, Donz{\'e}, Maasoumy, Murray, Sangiovanni-Vincentelli, and Seshia]{raman2014model}
Vasumathi Raman, Alexandre Donz{\'e}, Mehdi Maasoumy, Richard~M Murray, Alberto Sangiovanni-Vincentelli, and Sanjit~A Seshia.
\newblock Model predictive control with signal temporal logic specifications.
\newblock In \emph{53rd IEEE Conference on Decision and Control}, pages 81--87. IEEE, 2014.

\bibitem[Sadraddini and Belta(2015)]{sadraddini2015robust}
Sadra Sadraddini and Calin Belta.
\newblock Robust temporal logic model predictive control.
\newblock In \emph{2015 53rd Annual Allerton Conference on Communication, Control, and Computing (Allerton)}, pages 772--779. IEEE, 2015.

\bibitem[Sakoe and Chiba(1978)]{sakoe1978dynamic}
Hiroaki Sakoe and Seibi Chiba.
\newblock Dynamic programming algorithm optimization for spoken word recognition.
\newblock \emph{IEEE transactions on acoustics, speech, and signal processing}, 26\penalty0 (1):\penalty0 43--49, 1978.

\bibitem[Schulman et~al.(2017)Schulman, Wolski, Dhariwal, Radford, and Klimov]{schulman2017proximal}
John Schulman, Filip Wolski, Prafulla Dhariwal, Alec Radford, and Oleg Klimov.
\newblock Proximal policy optimization algorithms.
\newblock \emph{arXiv preprint arXiv:1707.06347}, 2017.

\bibitem[Sreedharan et~al.(2020)Sreedharan, Srivastava, and Kambhampati]{sreedharan2020tldr}
Sarath Sreedharan, Siddharth Srivastava, and Subbarao Kambhampati.
\newblock Tldr: Policy summarization for factored ssp problems using temporal abstractions.
\newblock In \emph{Proceedings of the International Conference on Automated Planning and Scheduling}, volume~30, pages 272--280, 2020.

\bibitem[Todorov et~al.(2012)Todorov, Erez, and Tassa]{todorov2012mujoco}
Emanuel Todorov, Tom Erez, and Yuval Tassa.
\newblock Mujoco: A physics engine for model-based control.
\newblock In \emph{2012 IEEE/RSJ international conference on intelligent robots and systems}, pages 5026--5033. IEEE, 2012.

\bibitem[Wang et~al.(2017)Wang, Merel, Reed, de~Freitas, Wayne, and Heess]{wang2017robust}
Ziyu Wang, Josh~S Merel, Scott~E Reed, Nando de~Freitas, Gregory Wayne, and Nicolas Heess.
\newblock Robust imitation of diverse behaviors.
\newblock \emph{Advances in Neural Information Processing Systems}, 30, 2017.

\bibitem[Xiang et~al.(1997)Xiang, Sun, Fan, and Gong]{xiang1997generalized}
Yang Xiang, DY~Sun, W~Fan, and XG~Gong.
\newblock Generalized simulated annealing algorithm and its application to the thomson model.
\newblock \emph{Physics Letters A}, 233\penalty0 (3):\penalty0 216--220, 1997.

\bibitem[Xu et~al.(2018)Xu, Saha, Hu, Mishra, and Julius]{xu2018advisory}
Zhe Xu, Sayan Saha, Botao Hu, Sandipan Mishra, and A~Agung Julius.
\newblock Advisory temporal logic inference and controller design for semiautonomous robots.
\newblock \emph{IEEE Transactions on Automation Science and Engineering}, 16\penalty0 (1):\penalty0 459--477, 2018.

\bibitem[Yaghoubi and Fainekos(2019)]{yaghoubi2019worst}
Shakiba Yaghoubi and Georgios Fainekos.
\newblock Worst-case satisfaction of stl specifications using feedforward neural network controllers: a lagrange multipliers approach.
\newblock \emph{ACM Transactions on Embedded Computing Systems (TECS)}, 18\penalty0 (5s):\penalty0 1--20, 2019.

\end{thebibliography}

\newpage
\appendix
\section{Decoding and Translating TLINet}\label{TLINet}
Consider a TLINet with five layers, as illustrated in Figure \ref{fig:NN}. In this example, the network begins with a predicate layer with two predicate modules, followed by a Boolean layer with two modules, then two temporal layers each with two modules, and finally a Boolean layer containing a single module. The parameters include spatial parameters $a$ and $b$, a selector $\kappa$ for the type of operator, temporal parameters $t_1$ and $t_2$, and Boolean parameter $w$ that determines module activation. For instance, in the final layer, $w=[0,1]$ indicates that the first module is deactivated and the second module is activated in this Boolean operation. The structure of the STL formula is determined by the activation of modules, guided by the learned Boolean parameters. Such a network can be succinctly translated into an STL formula $\psi = \always_{[0,10]}\event_{[3,7]}((x>0.9)\land(x<-0.7))$.
\begin{figure}[h]
\subfigure[Initialize TLINet structure.]{
\begin{tikzpicture}
\begin{scope}[every node/.style={minimum width=2cm,draw}]
\node[draw=DodgerBlue3] (layer 1a) {$(a,b)$};
\node[draw=DodgerBlue3,right= 3mm of layer 1a] (layer 1b) {$(a,b)$};
\node[draw=black!30!green,below=5mm of layer 1a] (layer 2a) {$(\kappa,w)$};
\node[draw=black!30!green] (layer 2b) at (layer 2a-|layer 1b) {$(\kappa,w)$};
\node[draw=orange,below=5mm of layer 2a] (layer 3a) {$(\kappa,t_1,t_2)$};
\node[draw=orange] (layer 3b) at (layer 3a-|layer 2b) {$(\kappa,t_1,t_2)$};
\node[draw=orange,below=5mm of layer 3a] (layer 4a) {$(\kappa,t_1,t_2)$};
\node[draw=orange] (layer 4b) at (layer 4a-|layer 3b) {$(\kappa,t_1,t_2)$};
\node[draw=black!30!green,below of=layer4a] (layer 5) at ($(layer 4a)!0.5!(layer 4b)$) {$(\kappa,w)$};
\end{scope}
\begin{scope}[every note/.style={minimum width=0.5cm}]
\node[DodgerBlue3,left=0mm of layer 1a,font=\fontsize{6}{6}\selectfont] {\textsf{Predicate}};
\node[black!30!green,left=0mm of layer 2a,font=\fontsize{6}{6}\selectfont] {\textsf{Boolean}};
\node[orange,left=0mm of layer 3a,font=\fontsize{6}{6}\selectfont] {\textsf{Temporal}};
\node[orange,left=0mm of layer 4a,font=\fontsize{6}{6}\selectfont] {\textsf{Temporal}};
\node[black!30!green,left=0mm of layer 5,font=\fontsize{6}{6}\selectfont] {\textsf{Boolean}};
\end{scope}
\node[circle,fill=orange!70,minimum size=0.8cm] at ($(layer 1a)!0.5!(layer 1b)+(0,1)$) (s) {$\mathbf{y}$};
\node[circle,fill=orange!70,minimum size=0.8cm, below=5mm of layer 5] (r) {$\tilde{r}$};
\begin{scope}[-latex]
\draw (s) -- (layer 1a);
\draw (s) -- (layer 1b);
\draw (layer 1a) -- (layer 2a);
\draw (layer 1a) -- (layer 2b);
\draw (layer 1b) -- (layer 2a);
\draw (layer 1b) -- (layer 2b);
\draw (layer 2a) -- (layer 3a);
\draw (layer 2b) -- (layer 3b);
\draw (layer 3a) -- (layer 4a);
\draw (layer 3b) -- (layer 4b);
\draw (layer 4a) -- (layer 5);
\draw (layer 4b) -- (layer 5);
\draw (layer 5) -- (r);
\end{scope}
\end{tikzpicture}
}
\subfigure[Learned TLINet parameters.]{
\begin{tikzpicture}
\begin{scope}[every node/.style={minimum width=2cm,draw}]
\node[draw=DodgerBlue3] (layer 1a) {$(1.0,0.9)$};
\node[draw=DodgerBlue3,right= 3mm of layer 1a] (layer 1b) {$(-1.0,0.7)$};
\node[draw=black!30!green,below=5mm of layer 1a] (layer 2a) {$(1,[1,1])$};
\node[draw=black!30!green] (layer 2b) at (layer 2a-|layer 1b) {$(0,[1,1])$};
\node[draw=orange,below=5mm of layer 2a] (layer 3a) {$(0,0,15)$};
\node[draw=orange] (layer 3b) at (layer 3a-|layer 2b) {$(1,3,7)$};
\node[draw=orange,below=5mm of layer 3a] (layer 4a) {$(1,5,10)$};
\node[draw=orange] (layer 4b) at (layer 4a-|layer 3b) {$(0,0,10)$};
\node[draw=black!30!green,below of=layer4a] (layer 5) at ($(layer 4a)!0.5!(layer 4b)$) {$(1,[0,1])$};
\end{scope}
\node[circle,fill=orange!70,minimum size=0.8cm] at ($(layer 1a)!0.5!(layer 1b)+(0,1)$) (s) {$\mathbf{y}$};
\node[circle,fill=orange!70,minimum size=0.8cm, below=5mm of layer 5] (r) {$\tilde{r}$};
\begin{scope}[-latex]
\draw (s) -- (layer 1a);
\draw (s) -- (layer 1b);
\draw (layer 1a) -- (layer 2a);
\draw (layer 1a) -- (layer 2b);
\draw (layer 1b) -- (layer 2a);
\draw (layer 1b) -- (layer 2b);
\draw (layer 2a) -- (layer 3a);
\draw (layer 2b) -- (layer 3b);
\draw (layer 3a) -- (layer 4a);
\draw (layer 3b) -- (layer 4b);
\draw (layer 4a) -- (layer 5);
\draw (layer 4b) -- (layer 5);
\draw (layer 5) -- (r);
\end{scope}
\end{tikzpicture}
}
\subfigure[TLINet modules to STL subformulas.]{
\begin{tikzpicture}
\begin{scope}[every node/.style={minimum width=2cm,draw}]
\node[draw=DodgerBlue3] (layer 1a) {$x>0.9$};
\node[draw=DodgerBlue3,right= 3mm of layer 1a] (layer 1b) {$x<-0.7$};
\node[draw=black!30!green,text width=1.3cm,,font=\fontsize{6}{6}\selectfont,below=5mm of layer 1a] (layer 2a) {$(x>0.9)\lor(x<-0.7)$};
\node[draw=black!30!green,text width=1.3cm,,font=\fontsize{6}{6}\selectfont] (layer 2b) at (layer 2a-|layer 1b) {$(x>0.9)\land(x<-0.7)$};
\node[draw=orange,below=5mm of layer 2a] (layer 3a) {$\phi_{11}=\always_{[0,15]}$};
\node[draw=orange] (layer 3b) at (layer 3a-|layer 2b) {$\phi_{12}=\event_{[3,7]}$};
\node[draw=orange,below=5mm of layer 3a] (layer 4a) {$\phi_{21}=\event_{[5,10]}$};
\node[draw=orange] (layer 4b) at (layer 4a-|layer 3b) {$\phi_{22}=\always_{[0,10]}$};
\node[draw=black!30!green,below of=layer4a] (layer 5) at ($(layer 4a)!0.5!(layer 4b)$) {$\psi=\phi_{22}\phi_{12}$};
\end{scope}
\node[circle,fill=orange!70,minimum size=0.8cm] at ($(layer 1a)!0.5!(layer 1b)+(0,1)$) (s) {$\mathbf{y}$};
\node[circle,fill=orange!70,minimum size=0.8cm, below=5mm of layer 5] (r) {$\tilde{r}$};
\begin{scope}[-latex]
\draw (s) -- (layer 1a);
\draw (s) -- (layer 1b);
\draw (layer 1a) -- (layer 2a);
\draw (layer 1a) -- (layer 2b);
\draw (layer 1b) -- (layer 2a);
\draw (layer 1b) -- (layer 2b);
\draw (layer 2a) -- (layer 3a);
\draw (layer 2b) -- (layer 3b);
\draw (layer 3a) -- (layer 4a);
\draw (layer 3b) -- (layer 4b);
\draw (layer 4a) -- (layer 5);
\draw (layer 4b) -- (layer 5);
\draw (layer 5) -- (r);
\end{scope}
\end{tikzpicture}
}
\caption{An example of a $5$-layer TLINet and how it can be transferred to an STL formula from learning parameters.}
\label{fig:NN}
\end{figure}

\section{Experiments Setup and Detailed Results} \label{appendix}
The inference network used in the Sec.~\ref{sec:navigation} and Sec.~\ref{sec:mujoco} consists of three layers: a predicate layer with $2n_f$ modules, a temporal layer with $2n_f$ modules, and a Boolean layer with $1$ modules, where $n_f$ is the dimension of the features. In the self-driving case in Sec.~\ref{sec:case-driving}, we observe improved performance with the addition of a second Boolean layer, so we use a similar structure with an extra Boolean layer. The policy network uses an LSTM~\cite{hochreiter1997long} with $3$ hidden layers, each containing $64$ nodes.

The STL formulas learned from different initializations are slightly different. In the following, we show one of them for each scenario. We also demonstrate some negative trajectories generated by the policies during the training to visualize the training processes.

\subsection{Random Navigation}
In Sec.~\ref{sec:navigation}, the features on which the STL formula is defined are the position of the robot and the distances from the robot to the centers of the regions, i.e., $y = h(x) = [p_x,p_y,d_{g_1},d_{g_2},d_o]$. The ground-truth STL formula that the expert policy is trained on is:
\begin{equation}
\begin{aligned}
    &\always_{[0,24]}(p_y<4)\land\always_{[0,24]}(p_y>0)\\
    \land&\event_{[0,8]}(d_{g_1}<0.5)\land \event_{[8,24]}(d_{g_2}<0.5) \land\always_{[0,24]}(d_o>0.5)
\end{aligned}
\end{equation}
A learned STL formula from one of the random initializations is given as below:
\begin{equation}
    \label{eq:stl-nav-learned}
    \begin{aligned}
        &\always_{[0,24]}(p_y<3.7) \land \always_{[1,17]}(p_y>0.41) \land \event_{[4,9]}(d_{g_1}<0.41) \\
        \land &\event_{[0,10]}(d_{g_1}<2.71) \land \event_{[8,24]}(d_{g_2}<0.48) \land \always_{[8,20]}(d_o>0.78),
    \end{aligned}
\end{equation}
which closely reproduces the specification the expert aims to satisfy.

\begin{figure}[t]
    \centering
    \subfigure[Iteration 0 (random)]{
    \centering
    \includegraphics[width=0.46\textwidth]{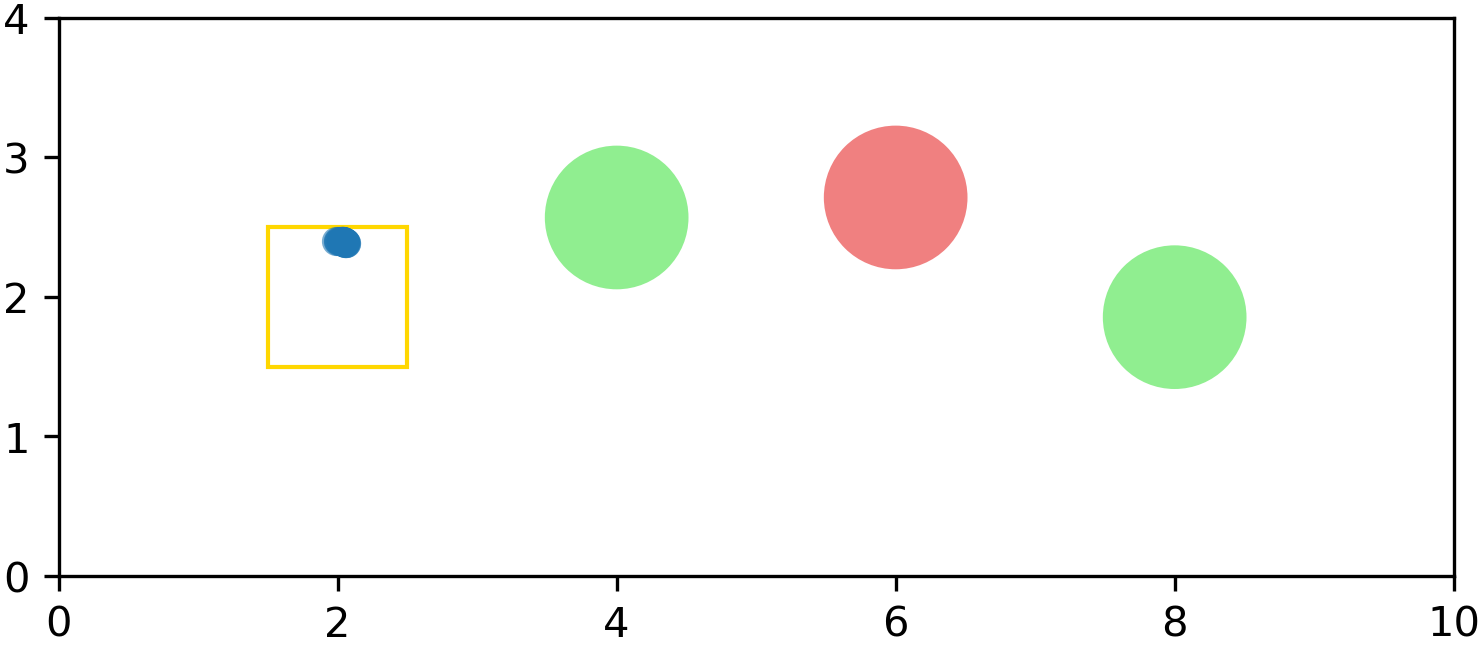}
    }
    \subfigure[Iteration 1]{
        \centering
        \includegraphics[width=0.46\textwidth]{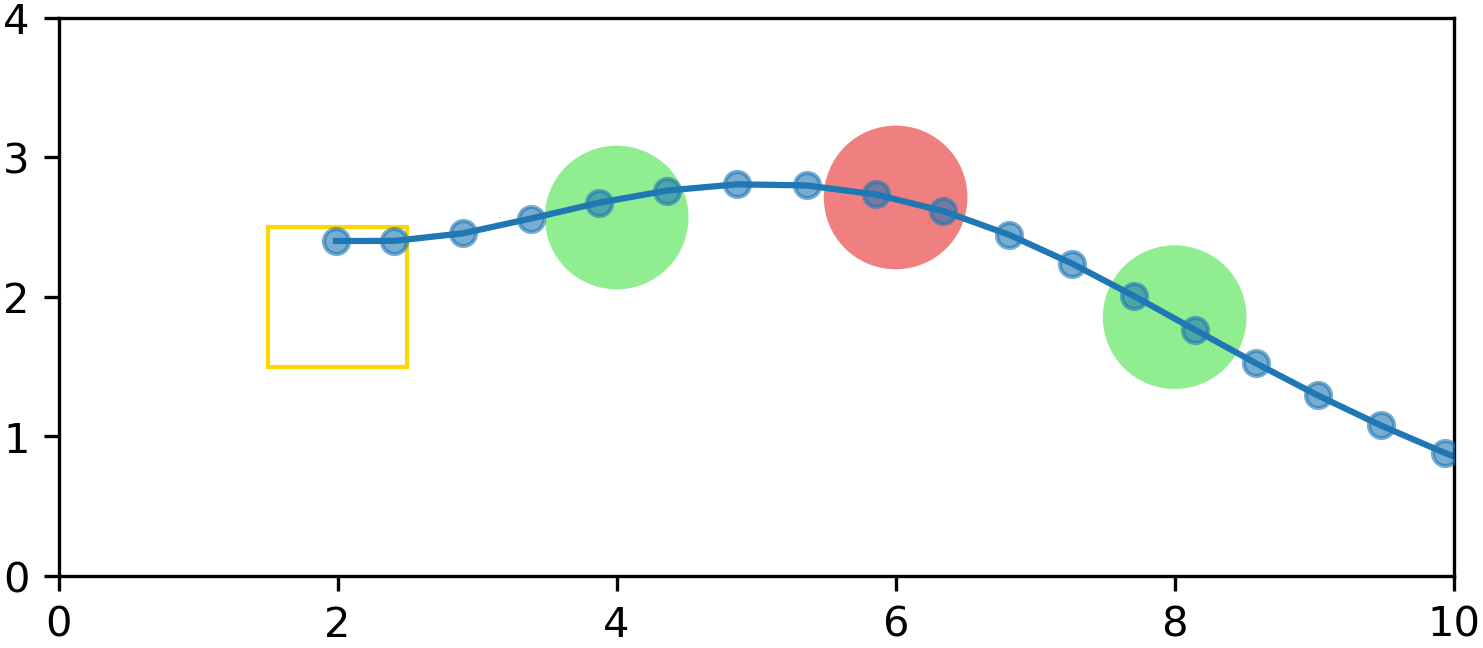}
    }
    \subfigure[Iteration 2]{
    \centering
    \includegraphics[width=0.46\textwidth]{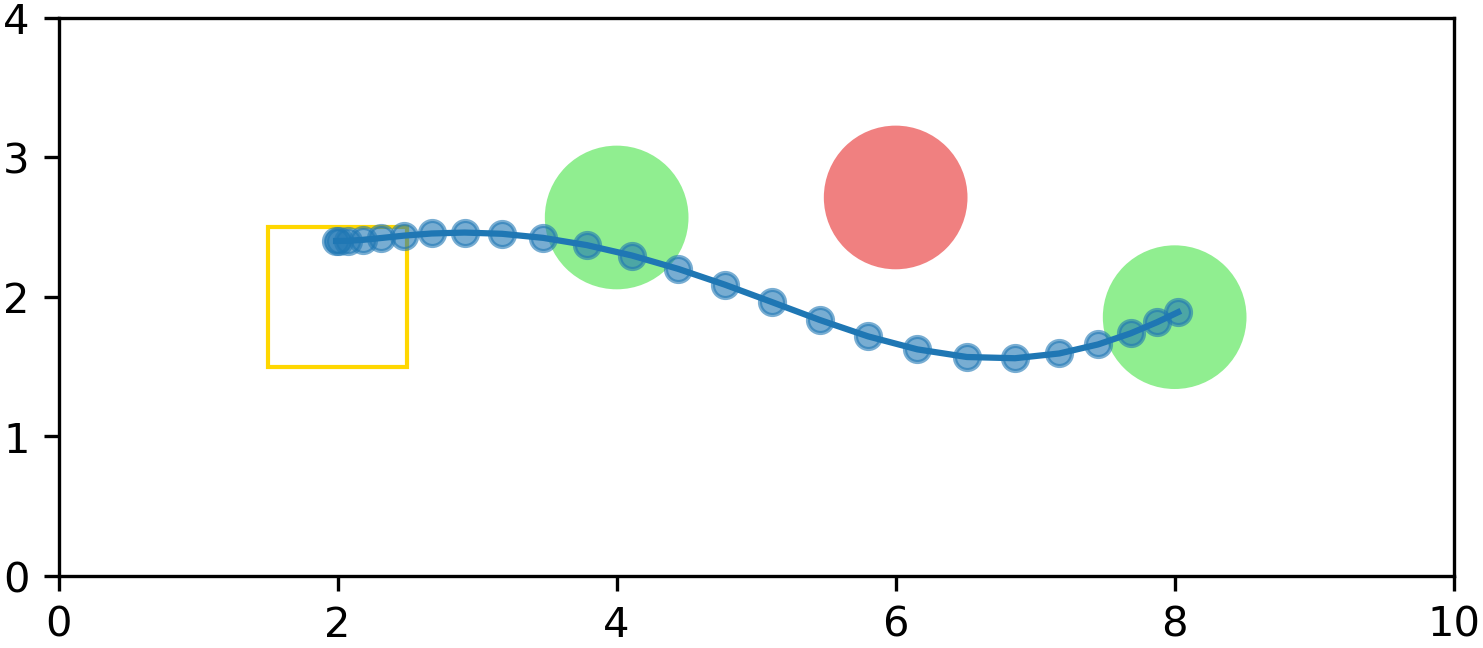}
    }
    \subfigure[Iteration 3]{
        \centering
        \includegraphics[width=0.46\textwidth]{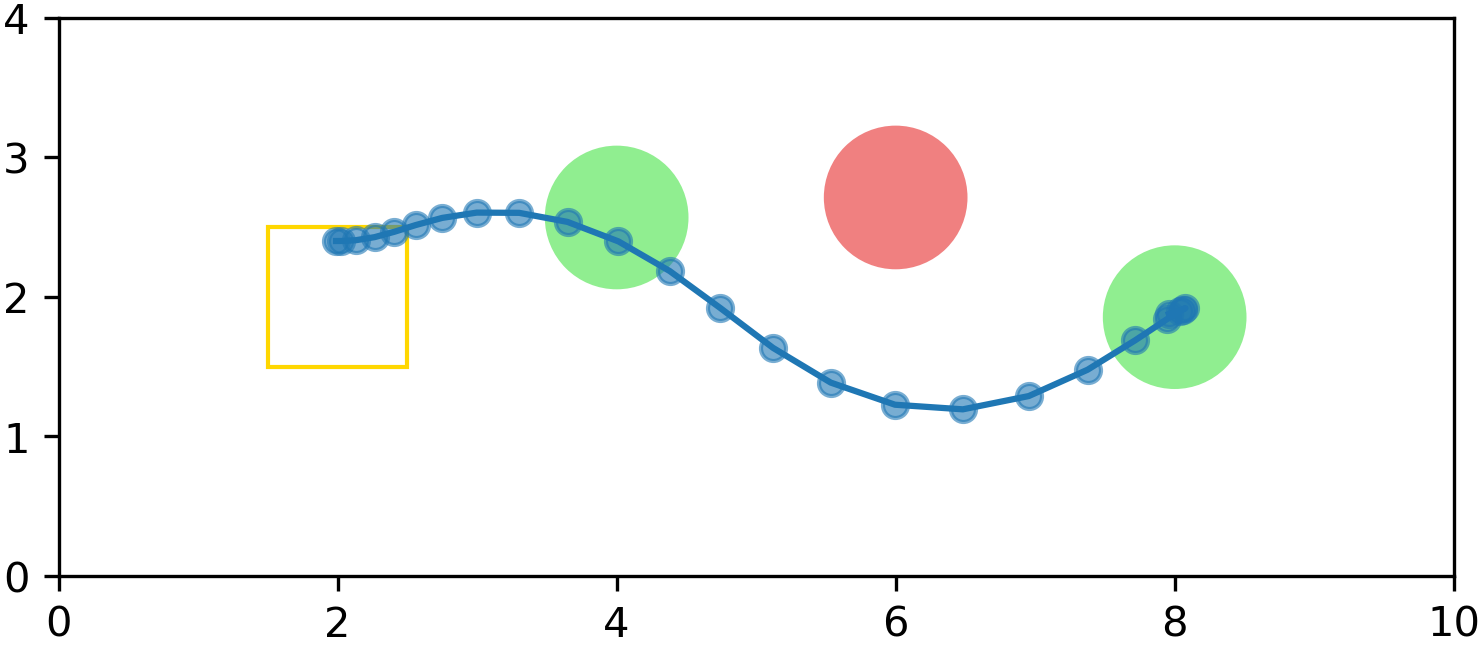}
    }
    \subfigure[Iteration 7]{
    \centering
    \includegraphics[width=0.46\textwidth]{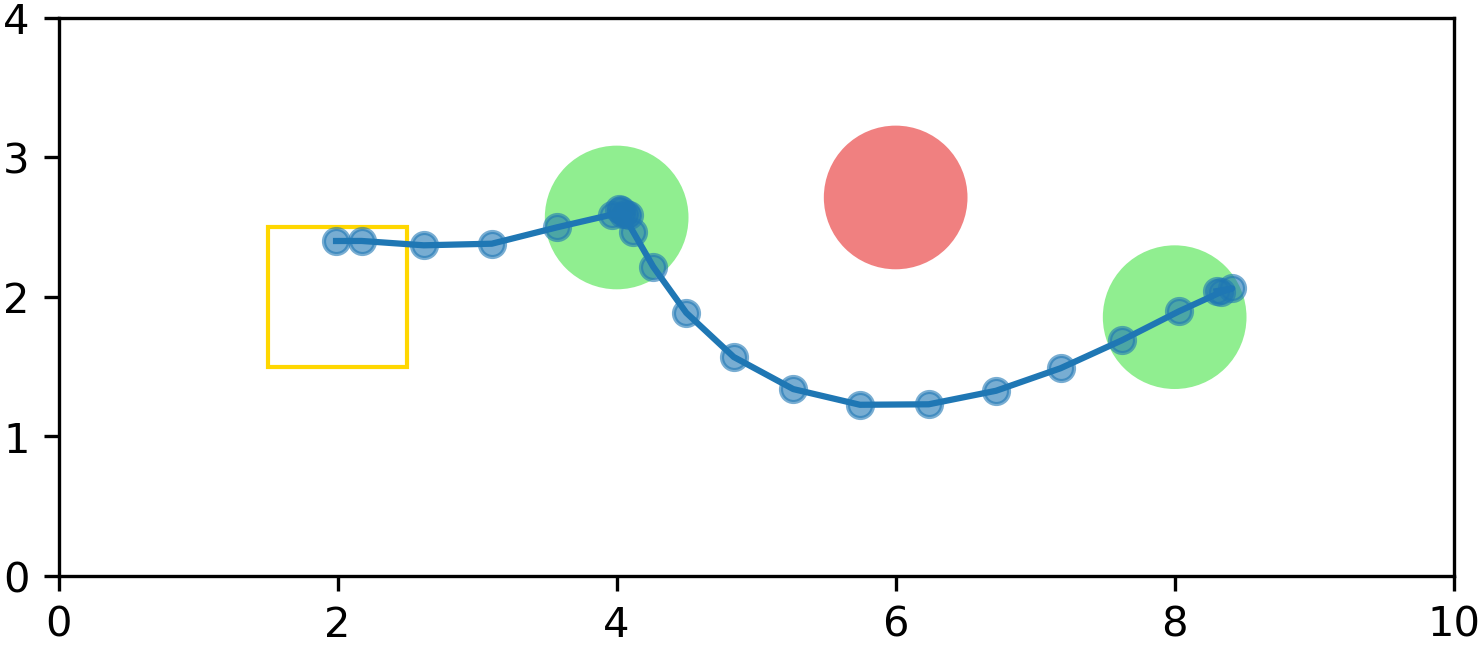}
    }
    \subfigure[Expert policy]{
        \centering
        \includegraphics[width=0.46\textwidth]{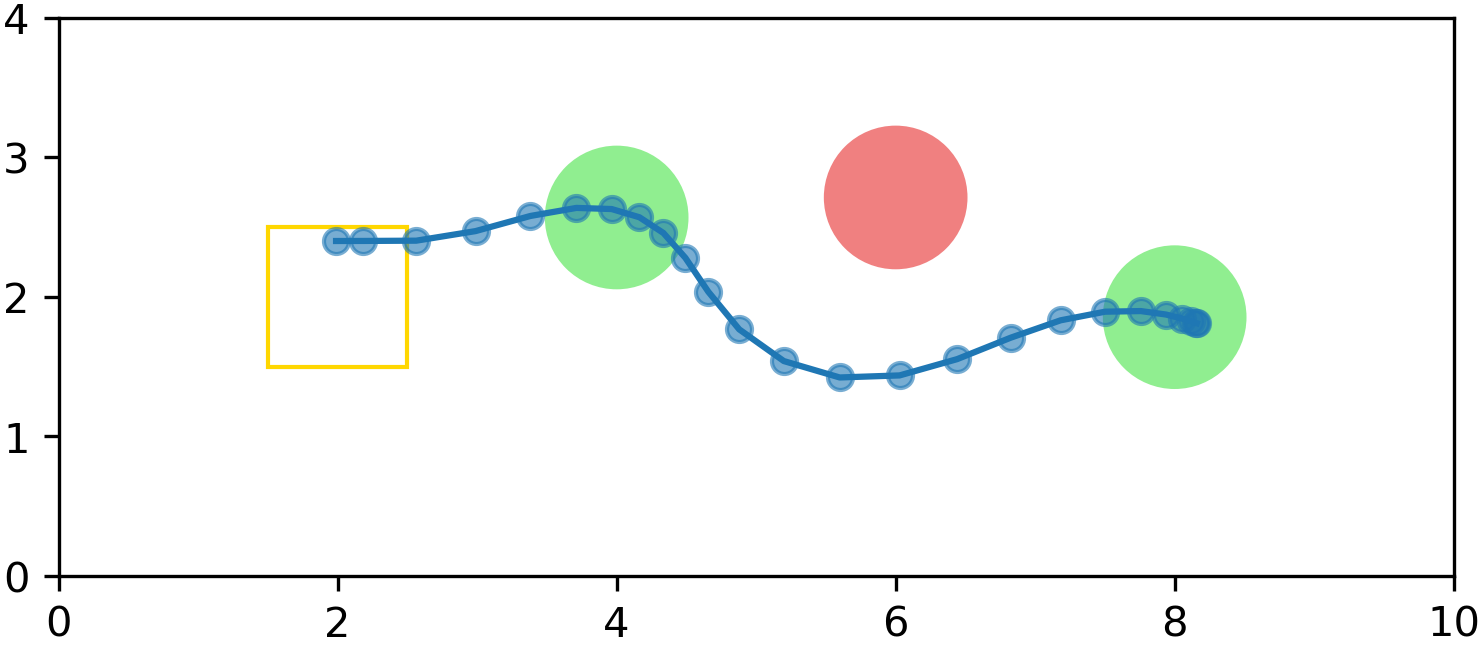}
    }
    \caption{Trajectories generated by the policy during the training process. }
    \label{fig:app-nav}
\end{figure}

Some negative trajectories generated by the policy during training under same initial and environment states are shown in Fig.~\ref{fig:app-nav}. At iteration 0, random trajectories are sampled and labeled as negative examples. These samples are insufficient to push the decision boundary near the expert demonstrations, resulting in a control policy that fails to replicate expert behavior. As training progresses and more informative negative examples are added, the policy-generated trajectories gradually converge toward the expert’s behavior.

\subsection{MuJoCo Tasks}
For the Inverted Pendulum in Sec.~\ref{sec:mujoco}, the features $y$ and system state $x$ are identical, both of them are same as the observation provided in OpenAI Gym~\cite{brockman2016openai}, i.e., $y = x = h(x) = [p,\theta,v,\omega]$, where $p$ and $v$ are the position and velocity of the cart along the linear surface, $\theta$ and $\omega$ are the angle and angular velocity of the pendulum. One of the learned STL formula is given as:
\begin{equation}
\label{eq:ip}
\begin{aligned}
    &\always_{[21,45]}(\theta<0.07) \land \always_{[23,32]}(v>-0.17) \\
    \land & \always_{[14,50]}(\omega>-0.79) \land \event_{[19,37]}(\omega>-0.16)
\end{aligned}
\end{equation}
The first subformula $\always_{[21,45]}(\theta<0.07)$ requires that the pendulum must not tilt in the positive direction, while the rest of the formula specifies additional constraints for the system. 

For the Reacher, the system state, including agent state and environment state, is same as the observation given by Gym: $$x = [\cos\theta_1, \cos\theta_2, \sin\theta_1, \sin\theta_2, p_x, p_y, \omega_1, \omega_2, d_x, d_y, d_z]\in\mathbb R^{11},$$ where $\theta_1$ and $\theta_2$ are the angles of the first and second arms, $\omega_1$ and $\omega_2$ are their angular velocities, $p_x$ and $p_y$ are the coordinates of the target, $d_x$, $d_y$ and $d_z$ are the differences between the target and the fingertip in the $x$, $y$, $z$ directions. The features we selected are $y = h(x) = [\cos\theta_1,\cos\theta_2,\sin\theta_1,\sin\theta_2,\omega_1,\omega_2,d_x,d_y]$.
One of the learned STL formula is:
\begin{equation}
\label{eq:reacher}
\begin{aligned}
    &\always_{[22,50]} (d_x < 0.04) \land \always_{[10,50]} (d_x > -0.04) \\
    \land &\event_{[16,23]} (d_y < 0.04) \land \always_{[15,28]} (d_y > -0.03) \\
    \land & \event_{[23,50]}(\sin{\theta_2} > 0.38) \land \event_{[19,37]}(\omega_2 < 0.4),
\end{aligned}
\end{equation}
The first $4$ subformulas in \eqref{eq:reacher} require the fingertip to reach the target, while the last $2$ subformulas describe how the expert accomplishes this as the inverse kinematics does not have a unique solution. 

\subsection{A Self-Driving Scenario}

In Sec.~\ref{sec:case-driving}, the features are identical with the system state, i.e., $y=h(x)=x$. The learned STL formula $\phi_{\theta_I}$ is:
\begin{equation}
\label{eq:case2-stl}
\begin{aligned}
    &\Big (\always_{[45,47]} (v_{ot}>2.35) \land \always_{[20,57]}(v_{eg}>1.31) \land \always_{[24,46]}(v_{eg}<5.55) \Big ) \lor \\
    &\Big (\event_{[42,55]}(v_{ot}<2.94) \land \event_{[20,57]}(v_{eg}<0.01) \land \always_{[24,46]}(v_{eg}<5.55) \Big ).
\end{aligned}
\end{equation}
Formula \eqref{eq:case2-stl} can be translated as 
``if $v_{ot}$ is always greater than $2.35$ over the time window $[45,47]$ (the other vehicle does not stop), ego's velocity should always be greater than $1.31$. If $v_{ot}$ eventually becomes less than $2.94$ in the time interval $[42,55]$ (the other vehicle decelerates), ego's velocity should also be eventually less than $0.01$ between $[20,57]$, which means a full stop". In both cases, $v_{eg}$ should be always less than $5.55$ in the time window $[24,46]$. This formula explicitly reflects the rules that the ego should obey.

\begin{figure}
    \centering
    \subfigure[\small Iteration 1]{
        \includegraphics[width=.3\textwidth]{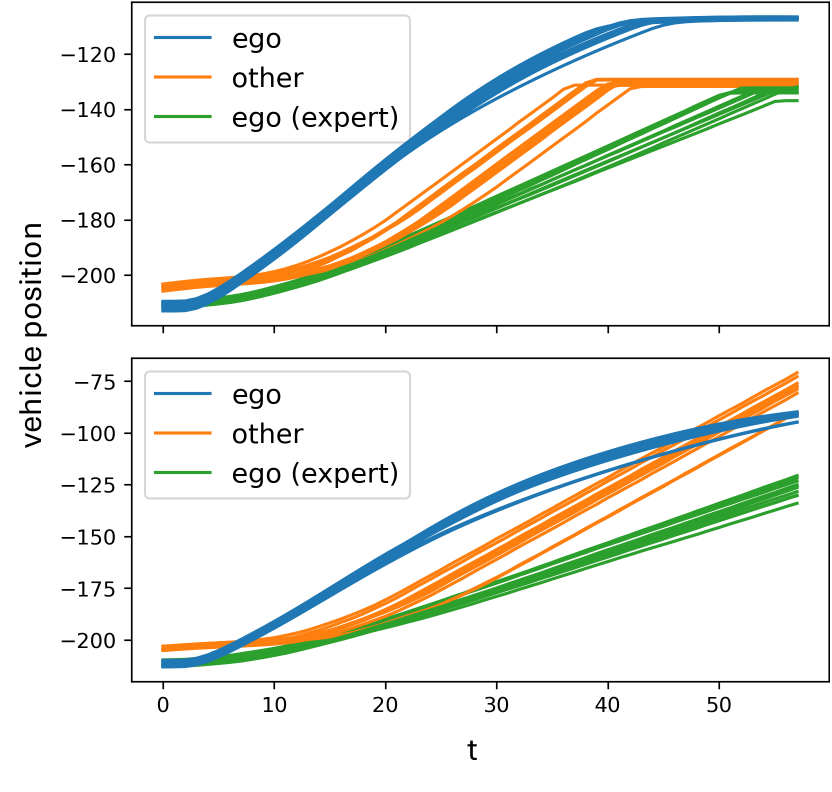}
    }
    \subfigure[\small Iteration 2]{
        \includegraphics[width=.3\textwidth]{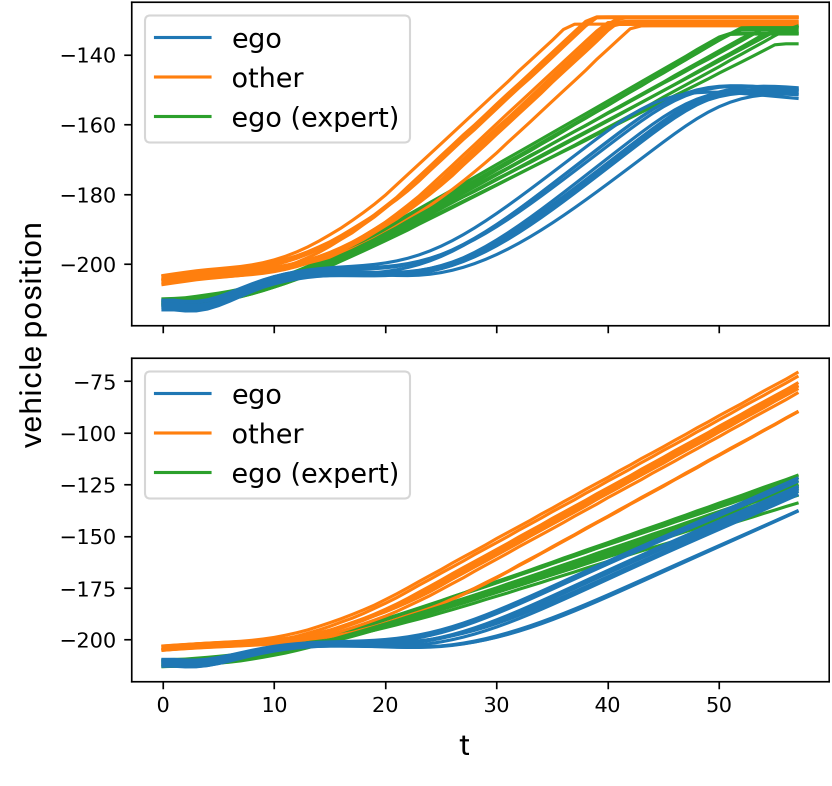}
    }
    \subfigure[\small Iteration 3]{
        \includegraphics[width=.3\textwidth]{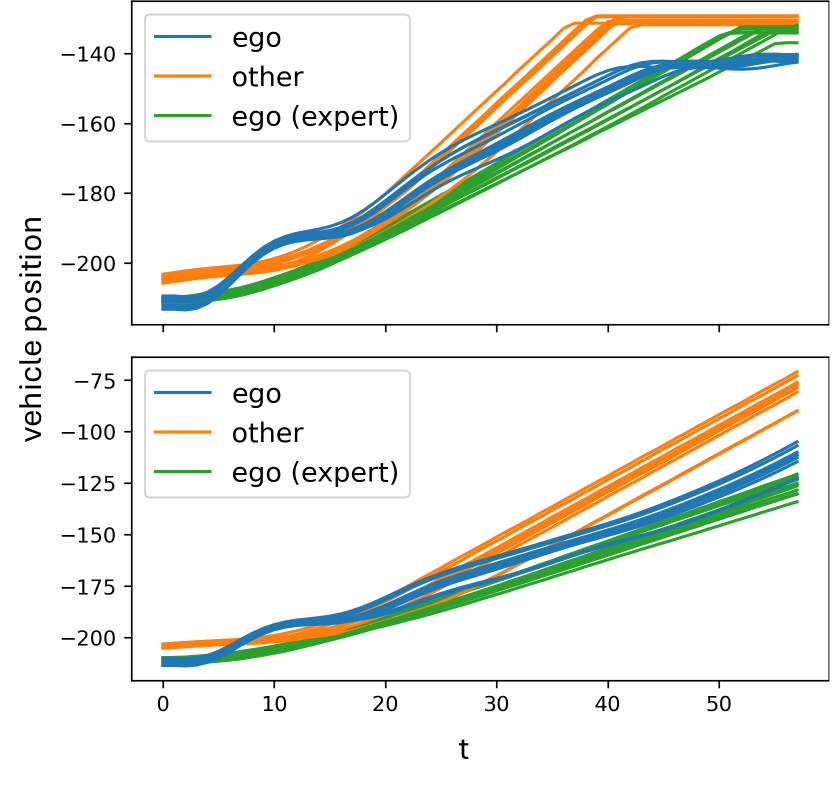}
    }
    \caption{(a)-(c): Sampled trajectories of ego and other vehicles' positions generated by the policy network after each iteration. y-axes are the positions of the ego and other vehicles. Blue curves stand for ego and orange curves stand for the other vehicle. Green curves are expert demonstrations of the ego. The top figures show the situations when there are pedestrians, while the bottom figures show the situations when there is no pedestrian.}
    \label{fig:app-driving}
\end{figure}

Some sampled trajectories of ego and other vehicles' positions generated by the learned policy during the training process are shown in Fig.~\ref{fig:app-driving}. Again, the trajectories increasingly align with the expert demonstrations, indicating progressive improvement in policy performance.

\end{document}